\documentclass[12pt]{article}

\usepackage{amsmath,amsthm, amsfonts, amssymb, amsxtra,amsopn}
\usepackage{pgfplots}
\usepgfplotslibrary{colormaps}
\pgfplotsset{compat=1.15}
\usepackage{pgfplotstable}
\usetikzlibrary{pgfplots.statistics} 
\usepackage{filecontents}
\usepackage{graphicx}
\usepackage{multirow}
\usepackage{tabu}
\usepackage{algorithmic}
\usepackage{longtable}
\usepackage{booktabs}
\usepackage{listings}
\usepackage{cmap}
\usepackage{colortbl}
\usepackage{adjustbox}
\usepackage{epsfig}
\usepackage{xstring}%%% needed for \rowvec macros

\usepackage{subfigure}

\PassOptionsToPackage{hyphens}{url}

\usepackage{hyperref}
\hypersetup{colorlinks=true,linkcolor=black,citecolor=black,urlcolor=blue,filecolor=black}
\hypersetup{pdfpagemode=UseNone,pdfstartview=}

\usepackage[tableposition=top]{caption}
\usepackage{enumitem}
\setlist[itemize]{noitemsep, topsep=0pt}

\advance\oddsidemargin by -0.45in
\advance\textwidth by 0.9in

\advance\topmargin by -0.3in
\advance\textheight by 0.6in

\long\def\symbolfootnotetext[#1]#2{\begingroup%
\def\thefootnote{\fnsymbol{footnote}}\footnotetext[#1]{#2}\endgroup}

\def\TP{\mbox{TP}}
\def\FP{\mbox{FP}}
\def\TN{\mbox{TN}}
\def\FN{\mbox{FN}}

\pgfkeys{
    /pgf/number format/fixed zerofill=true }
    
\pgfplotstableset{
    color cells/.code={%
        \pgfqkeys{/color cells}{#1}%
        \pgfkeysalso{%
            postproc cell content/.code={%
                \begingroup
                \pgfkeysgetvalue{/pgfplots/table/@preprocessed cell content}\value
\ifx\value\empty
\endgroup
\else
                \pgfmathfloatparsenumber{\value}%
                \pgfmathfloattofixed{\pgfmathresult}%
                \let\value=\pgfmathresult
                \pgfplotscolormapaccess
                    [\pgfkeysvalueof{/color cells/min}:\pgfkeysvalueof{/color cells/max}]%
                    {\value}%
                    {\pgfkeysvalueof{/pgfplots/colormap name}}%
                \pgfkeysgetvalue{/pgfplots/table/@cell content}\typesetvalue
                \pgfkeysgetvalue{/color cells/textcolor}\textcolorvalue
                \toks0=\expandafter{\typesetvalue}%
                \xdef\temp{%
                    \noexpand\pgfkeysalso{%
                        @cell content={%
                            \noexpand\cellcolor[rgb]{\pgfmathresult}%
                            \noexpand\definecolor{mapped color}{rgb}{\pgfmathresult}%
                            \ifx\textcolorvalue\empty
                            \else
                                \noexpand\color{\textcolorvalue}%
                            \fi
                            \the\toks0 %
                        }%
                    }%
                }%
                \endgroup
                \temp
\fi
            }%
        }%
    }
}

\def\srowvecc#1#2{(\!\begin{array}{cc} 
      \noexpandarg\IfBeginWith{#1}{-}{\! #1}{#1}
    & #2\kern-0.5pt\end{array}\!)}
\def\rowvecc#1#2{\left(\!\begin{array}{cc} 
      \noexpandarg\IfBeginWith{#1}{-}{\! #1}{#1}
    & #2\kern-0.5pt\end{array}\!\right)}
\def\rowveccc#1#2#3{\left(\!\begin{array}{ccc} 
      \noexpandarg\IfBeginWith{#1}{-}{\! #1}{#1}
    & #2 
    & #3\kern-0.5pt\end{array}\!\right)}
\def\rowvecccc#1#2#3#4{\left(\!\begin{array}{cccc}
      \noexpandarg\IfBeginWith{#1}{-}{\! #1}{#1}
    & #2 
    & #3 
    & #4\kern-0.5pt\end{array}\!\right)}
\def\srowvecccc#1#2#3#4{\bigl(\!\begin{array}{cccc}
      \noexpandarg\IfBeginWith{#1}{-}{\! #1}{#1}
    & #2 
    & #3 
    & #4\kern-0.5pt\end{array}\!\bigr)}
\def\rowveccccc#1#2#3#4#5{\left(\!\begin{array}{ccccc} 
      \noexpandarg\IfBeginWith{#1}{-}{\! #1}{#1}
    & #2
    & #3
    & #4
    & #5\kern-0.5pt\end{array}\!\right)}
\def\srowvecccccc#1#2#3#4#5#6{(\!\begin{array}{cccccc} 
      \noexpandarg\IfBeginWith{#1}{-}{\! #1}{#1}
    & #2
    & #3
    & #4
    & #5
    & #6\kern-0.5pt\end{array}\!)}
\def\rowvecccccc#1#2#3#4#5#6{\left(\!\begin{array}{cccccc} 
      \noexpandarg\IfBeginWith{#1}{-}{\! #1}{#1}
    & #2
    & #3
    & #4
    & #5
    & #6\kern-0.5pt\end{array}\!\right)}

\title{Free-Text Keystroke Dynamics for User Authentication}

\author{Jianwei Li\footnotemark[1]\ \ \ 
Han-Chih Chang\footnotemark[1]\ \ \ 
Mark Stamp\footnotemark[1]\,\,\footnotemark[2]}

\begin{document}

\symbolfootnotetext[1]{Department of Computer Science, San Jose State University}
\symbolfootnotetext[2]{mark.stamp$@$sjsu.edu}

\maketitle

\abstract
In this research, we consider the problem of verifying user identity based on
keystroke dynamics obtained from free-text. 
We employ a novel feature engineering 
method that generates image-like transition matrices. 
For this image-like feature, 
a convolution neural network (CNN) with cutout achieves the best results.
A hybrid model 
consisting of a CNN and a recurrent neural network (RNN) is
also shown to outperform previous research in this field.

\section{Introduction}\label{chap:introduction}

User authentication is a critically important task in cybersecurity. 
Password based authentication is widely used, as are various biometrics.
Examples of popular biometrics include fingerprint, facial recognition, and iris scan.
However, all of these authentication methods suffer from some problems. 
For example, passwords can often be guessed and are sometimes stolen, 
and most biometric systems require special hardware~\cite{a,f,1}. 
Moreover, research has shown, for example, that the accuracy of face and fingerprint 
recognition on the elderly is lower than for young people~\cite{z}. Thus, an
authentication method that can resolve some of these issues is desirable. 

Intuitively, it would seem to be difficult to mimic
someone's typing behavior to a high degree of precision.
Thus, patterns hidden in typing behavior in the form of keystroke dynamics 
might serve as a strong biometric.
One advantage of a keystroke dynamics based authentication scheme
is that it requires no specialized hardware. In addition, such a scheme can 
provide a non-intrusive means of continuous or ongoing authentication, which
can be viewed as a form or intrusion detection.
Coursera, an online learning website, currently employs typing characteristics 
as part of its login system~\cite{0}.

Research into keystroke dynamics began about~20 years ago~\cite{16}.
However, early results in this field were not impressive. 
Most of the existing research in keystroke dynamics has focused on fixed-text typing behavior, 
which is viewed as one-time authentication~\cite{2,15,e,a,1}. 
Compared with fixed-text keystroke dynamics, the free-text case presents some
additional challenges. First, the number of useful features may differ among 
input sequences. Second, the optimal length of a keystroke sequence for analysis is
a factor that must be considered---a longer sequence is slower to process and
might include more noise,
while a shorter sequence may lack sufficient distinguishing characteristics. 
Moreover, for free-text keystroke sequences, it is more challenging to extract an effective
pattern, thus the robustness of any solution is a concern.

In this paper, we consider the free-text keystroke dynamics-based 
authentication problem. For this problem, we propose and analyze a unique feature 
engineering technique. Specifically, we organize features into an image-like transition matrix 
with multiple channels, where each row and column represents a key on the keyboard, 
with the depth corresponding to different categories of features. Then a convolutional
neural network (CNN) model with cutout regularization is trained on this engineered
feature. To better capture the sequential nature of the problem, we also consider a hybrid model using 
our CNN approach in combination with a gated recurrent unit (GRU) network. 
We evaluate these two models on open free-text keystroke datasets and 
compare the results with previous work. We carefully consider the effect of 
different lengths of keystroke sequences and other parameters on the 
performance of our models.

The contribution of this paper include the following:
\begin{itemize}
\item A new feature engineering method that organizes features as an image-like matrix 
for free-text keystroke dynamics-based authentication.
\item An analysis of cutout regularization as a step in the image analysis process.
\item A careful analysis of various hyperparameters, including the length of keystroke sequence
in our models. 
\end{itemize}

The remainder of this paper is organized as follows. 
Section~\ref{chap:introduction} introduces the basic concept of keystroke dynamics-based 
authentication, and we outline our general approach to the problem. 
In Section~\ref{chap:background}, we discuss background topics,
including the learning techniques employed and the datasets we have used.
Section~\ref{chap:background} also provides a discussion of relevant previous work.
Section~\ref{chap:features} describes the features that we use and, in particular,
we discuss the feature engineering strategy that we employ to prepare the input data 
for our continuous classification models. Then, in Section~\ref{chap:architeture},
we elaborate on the architectures of the various models considered in this paper,
and we discuss the hyperparameter tuning process. Section~\ref{chap:result}
includes our experiments and analysis of the results. Finally, Section~\ref{chap:conclusion} 
provides a conclusion and points to possible directions for future work.

\section{Background}\label{chap:background}

Authentication is the process that allows a machine to verify the identity of a user.
By the nature of the problem, authentication is a classification task.  
Keystroke dynamics is one of many techniques that have been considered for 
authentication. One advantage of keystroke dynamics is that
such an approach requires no special hardware.  

Precision and recall are two metrics used to evaluate classification models. 
Precision is the fraction of true positive instances among those classified as
positive, while recall is the fraction of true positive instances that are correctly
classified as such. Table~\ref{categories:user-authentication} lists some 
examples of use cases, along with the general degree of precision and recall
that typically must be attained in a useful system. 
Depending on the scenario, too many false positives (i.e., low precision) 
can render an IDS impractical, 
but an IDS must detect intrusions (high recall) or it has
clearly failed to perform adequately. On the other hand, in the identification
problem, we must be confident that our identification is correct
(high precision), even if we fail to identify subjects in a number of cases (low recall)

\begin{table}[!htb]
\caption{Use cases for keystroke dynamics-based systems}\label{categories:user-authentication}
\centering
\adjustbox{scale=0.85}{
\begin{tabular}{c|cccc} \midrule\midrule
Text & Scenario & Precision & Recall & Input length\\ \midrule
Fixed & One-time Authentication & high & high & short \\
Free & Intrusive Detection & low  & high  & long \\
Either & Identification & high & low & either  \\ \midrule\midrule
\end{tabular}
}
\end{table}

We note in passing that
even if the precision and recall are both high, practical usage scenarios 
for keystroke dynamics based systems may be limited by the length of the keystroke 
sequence required for analysis. In cases where a short keystroke sequence suffices, 
the technique will be more widely applicable. 

For a usage scenario, consider password-protected user accounts.
Keystroke dynamics would provide a second line of defense in such
an authentication system. In a two-factor authentication system,
an attacker would need to also accurately mimic a users tying habits. Note that
the second ``factor'' (i.e., keystroke dynamics) is transparent from a user's perspective---the
keystroke-related biometric information is collected passively, and requires no additional actions
from a user beyond typing his or her password.

Even in cases where the length of the keystroke sequence must be relatively long in order
to achieve the necessary accuracy, keystroke dynamics systems
could still be useful. For example, suppose that a user needs to reset their password 
for a high-security application, such as an online bank account. 
Most such systems require the user to answer a ``secret''
security question or multiple such questions. 
It can be difficult for users to remember the answers to security
questions, and the answers themselves (e.g., ``mother's maiden name'') are often not secret.
Replacing these question with a keystroke dynamics system would
free the user from the need to remember answers, as the user would 
simply need to type a sufficient number of characters in the user's
usual typing mode. 

From the use-case point of view, keystroke dynamics-based systems can 
be classified into those for which long input sequences are acceptable,
and those for which short input sequences are essential. 
We can also classify keystroke dynamics systems according to whether
they are based on fixed-text or free-text. In this paper, we only consider free-text.

\subsection{Related Work}

Previously, most work in keystroke dynamics was based on fixed-text,
but recently more attention has been paid to free-text keystroke analysis. 
There are two commonly used free-text keystroke datasets, which 
we refer to as the Buffalo dataset~\cite{i} and the Clarkson~II dataset~\cite{clark}. 
We discuss these datasets in more detail in Section~\ref{sec:datasets}.
Yan et al.~\cite{i} introduced the Buffalo dataset, which they use to 
evaluate a Gaussian mixture model (GMM) proposed by Hayreddin et al.~\cite{14}. 
The best EER obtained is~0.01. Their experiments are limited to 
keystroke data generated using the same keyboard. In our research, 
we evaluate our models on the entire Buffalo dataset, which includes different keyboards.

Pilsung et al.~\cite{12} divide the keyboard into three areas,  left, right, and space,
which correspond to the keys that are typically typed by the left hand (L), 
right hand (R), and thumbs (S), respectively. In this way, the time-based features extracted from 
different adjacent keystroke pairs fall into eight categories, which are denoted as
L-L, L-R, R-R, R-L, R-S, S-R, L-S, S-L. Then they compute average time-based 
histogram over each group and concatenate these values to form a feature vector. 
In this way, the free-text keystroke sequence is embedded into a vector of fixed length eight, 
which can then be used in different detection models. However, their method fails to 
preserve most of the sequential information that is available in keystrokes. 

To improve the performance of authentication systems based on free-text, 
Junhong et al.~\cite{b} propose a novel user-adaptive feature extraction method 
to capture unique typing pattern behind keystroke sequences. The method
consists of ranking time-based features, and splitting
all of these features into eight categories based on the rank order. 
Similar to the method proposed in~\cite{12}, they calculate the average time-based 
feature of each category as a single feature value and concatenate these features 
to form a vector. Their experiments show that the method significantly improves
performance, as compared with the method in~\cite{12}. 
However, they are still discarding a significant amount of the information
available in the raw keystroke dynamics data.

Eduard et al.~\cite{f} explore the use of multi-layer perceptrons (MLP) 
for keystroke based authentication. Their model considers time-based information 
between different keys separately, and does not aggregate information 
from the entire keystroke sequence. The performance appears to be relatively poor.

Mario et al.~\cite{d} propose a feature extraction model to capture user
input patterns. They test the impact of different numbers of layers in
various deep learning networks and compared the effectiveness of deep networks 
with classical machine learning methods. They attain a highest accuracy of~99.9\%\ 
using an MLP with nine hidden layers. However, their architectures 
are limited to feed-forward fully-connected layers, and better results
require a large number of hidden layers. Also, the dataset used in their research is different
from that used in our research, and thus the results are not directly comparable.

Kobojek et al.~\cite{u} uses an RNN-based model for classification based on keystroke data. 
They make use of keystroke sequential data. They achieve a best EER 
that is relatively high at~13.6\%.

Influenced by the work in~\cite{u},
Xiaofeng et al.~\cite{a} divide continuous keystroke dynamics sequences into keystroke 
subsequences of a fixed length and extracts time-based features from each subsequence. 
These features are then organized into a fixed-length sequence, and the 
resulting data is fed to a complex model consisting of a combined CNN and RNN. 
They consider an overlapping sliding window, and the they
use a majority vote system to further improve the accuracy. 
They best EERs of~2.67\%\ and~6.61\%\ over a pair of open free-text keystroke datasets. 
In our research, we propose a new architecture that is inspired by the model in~\cite{a}.

\subsection{Datasets}\label{sec:datasets}

In this paper, we evaluate various models based on two open-source 
free-text keystroke dynamics datasets. 
The two datasets we consider are from Clarkson University and 
the SUNY Buffalo. Next, we discuss these datasets.

\subsubsection{Buffalo Keystroke Dataset}

The Buffalo free-text keystroke dataset was collected 
by researchers at SUN Buffalo from~148 research subjects.
In this dataset, the subjects were asked to finish two typing tasks in a laboratory. 
For the first task, participants transcribed Steve Jobs' Stanford 
commencement speech, which was split into three parts. The second task 
consisted of responses to several free-text questions. 
The interval between the two sessions was~28 days. 
Additionally, only~75 of the subjects completed both typing tasks with 
the same keyboard, while the remaining~73 subjects typed using
three different keyboards across three sessions.

The Buffalo dataset includes relatively limited information. Specifically, the
the key that was pressed, along with timestamps for the key-down time and  
the key-up events. The average number of keystrokes in the three sessions 
exceeded~17,000 for each subject. 
Additionally, some of the participants used keyboards with different key 
layouts to input text information.
This dataset also provides gender information for each subject.

\subsubsection{Clarkson~II Keystroke Dataset}

The Clarkson~II keystroke dataset is a popular free-text keystroke dynamics dataset
that was collected by researchers at Clarkson University. This dataset includes keystroke timing 
information for~101 subjects in a completely uncontrolled and natural setting,
with the date having been collected over a period of~2.5 years. Compared with other 
datasets which are controlled to some degree, the participants 
contribute their data with different computers, different keyboards, different browsers, 
different software, and even different tasks (e.g., gaming, email, etc.). 
Models that perform well on this dataset should also perform well in a real-world scenario.

Unfortunately, the Clarkson~II dataset only provides 
very limited features---specifically, the timestamps of key-down and key-up events. 
The average number of keystrokes for each research subject is about~125,000. 
However, the number of keystroke events is far from uniform,
with some users having contributing only a small number of keystrokes. 
Therefore, we set a threshold of~20,000 keystrokes, which gives us
only~80 subjects.

\subsection{Deep Leaning Algorithms}

In this research, we apply deep learning methods to the free-text keystroke datasets discussed above. 
Our best-performing architecture is a novel combination of neural network based techniques.
In this section, we briefly discuss the learning techniques that we have employed.

\subsubsection{Multilayer Perceptron}

Multilayer perceptrons (MLP)~\cite{8} are a class of supervised learning algorithms 
with at least one hidden layer. 
Any MLP consists of a collection of interconnected artificial neurons, which are loosely modeled after
the neurons in the human brain. Nonlinearity is provided by the choice of the activation function 
in each layer. MLP is related to the classic machine learning technique of support vector machines (SVM).

\subsubsection{Convolutional Neural Network}

Convolutional neural networks (CNN)~\cite{9} are a special class of neural networks that 
make use of convolutional kernels to efficiently deal with local structure. CNNs are often ideal
for applications where local structure dominates, such as image analysis. CNNs with
multiple convolutional layers are able to extract the semantic information at different resolutions 
and have proven to be extremely powerful in computer vision tasks.

\subsubsection{Recurrent Neural Network}

Recurrent neural networks (RNN)~\cite{10} are used to deal with sequential or time-series data. 
For example, sequential information is essential for the analysis of text and speech. 
Plain ``vanilla'' RNNs suffer from vanishing gradients and related pathologies. To overcome these
issues, highly specialized RNN architectures have been developed, including
long short-term memory (LSTM)~\cite{10} and gated recurrent units (GRU)~\cite{19}. 
In practice, LSTMs and GRUs are among the most successful architectures yet developed.
In this research, we focus on GRUs, which are faster to train than LSTMs, and perform
well in our application.

\subsubsection{Cutout}\label{sec:cutout}

Fully connected neural networks often employ dropouts~\cite{17} to reduce 
overfitting problems. While dropouts work well for models with fully-connected layers,
the technique is not suitable for CNNs. Instead, we use cutout regularization~\cite{11} 
with our CNN models.
Cutouts are essentially the image-based equivalent of dropouts---we cut out part of the
image when training, which forces the CNN to learn from other parts of the image.
In addition to helping with overfitting, a model that is able to handle 
images with such occlusions is likely to be more robust. 

\section{Feature Engineering}\label{chap:features}

As mentioned in Section~\ref{chap:background}, we consider two open source keystroke datasets. 
Both the Buffalo and Clarkson~II datasets are free-text, and only provide 
fairly limited information. Therefore, we will have to consider feature engineering as a critical
part of our experiments. In this section we consider different categories 
of features and various types of feature engineering. 

\subsection{Features}\label{sect:features}

With the development of mobile devices, modern keyboards are no longer limited 
to physical keyboards, but also include most virtual devices that allow user input. 
Ideally, we would like to consider patterns in user typing behavior with respect to time-based information
and pressure-based features. However, pressure-based features are not directly available
from the datasets used in this research. In the future, datasets obtained using mobile
devices could include such information, which should enable stronger authentication
and identification results.

\begin{figure}[!htb]
\centering
\includegraphics[width=0.8\textwidth]{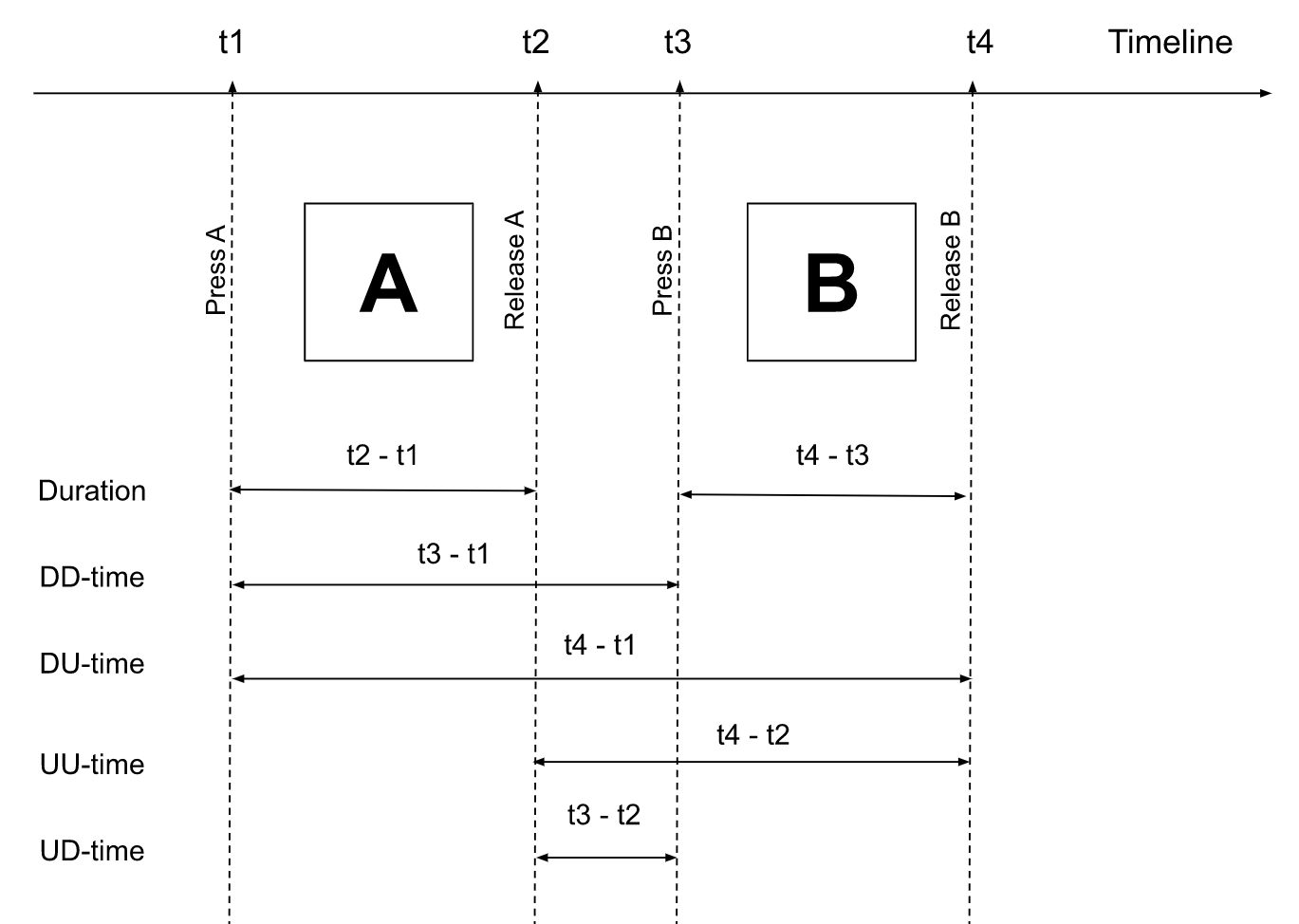}
\caption{Five time-based features} 
\label{fig:time_based_features}
\end{figure}

Again, in this research we necessarily focus on time-based features, 
because that is what we have available in our keystroke datasets. 
The five time-based features that we consider are illustrated 
in Figure~\ref{fig:time_based_features}.

Let~\texttt{A} and~\texttt{B} represent two consecutive input keys, 
with press and release representing a key-down 
and key-up event, respectively. 
The five time-based features are duration, down-down time (DD-time), 
up-down time (UD-time), up-up time (UU-time), and down-up time (DU-time). 
Duration is the time that the user holds a key in the down position,
while the other four features are clear from the figure. 
Note that for any two consecutive keystroke events, say, \texttt{A} and \texttt{B},
six features can be extracted, namely,
duration-\texttt{A}, 
duration-\texttt{B}, 
DD-time, 
UD-time, 
UU-time, and 
DU-time. 

\subsection{Length of Keystroke Sequence}

As mentioned in Section~\ref{chap:background}, we can divide keystroke dynamics-based authentication 
into four categories depending on the length and consistency of the keystroke sequence. 
For our free-text keystroke datasets, 
the data consists of a long keystroke sequence of thousands of characters
for each user. In previous research, such long sequences have been split 
into multiple subsequences, and we do the same here. Each subsequence is viewed 
as an independent keystroke sequence from the corresponding user. Previous research 
has shown that short keystroke subsequences decrease accuracy, while the longer keystroke 
subsequences may incorporate more noise. Therefore, we will need to 
experiment with different lengths of keystroke subsequence to determine 
an optimal value.

\subsection{Keystroke Dynamics Image}\label{sec:kimage}

In Section~\ref{chap:background}, we introduced the keystroke datasets used in this paper. 
As mentioned in the previous section, we divide the entire keystroke sequence into multiple 
subsequences, and in Section~\ref{sect:features} we discussed the six types of timing features
that are available. Thus, for a subsequence of length~$N$, there are~$6 (N-1)$
features that can be determined from consecutive pairs of keystrokes,
where repeated pairs are averaged and treated as a single pair.
For example, for a subsequence of length~50, we obtain
at most~$6\cdot 49 = 294$ features.
We view each keystroke subsequence as an independent input sequence for the
corresponding user. 
Next, we propose a new feature engineering structure to better organize these features. 

The features UD-time, DD-time, DU-time, and UU-time are determined by consecutive 
keystroke events. Therefore, we organize these four features into a transition matrix with four channels,
which can be viewed as four~$N\times N$ matrices overlaid. This approach is inspired by
RGB images, which have a depth of three, due to the~R, G, and~B channels.

\begin{figure}[!htb]
\centering
\begin{tikzpicture}[scale=0.27, every node/.style={scale=0.725}]
%
%%% layer 5
\draw[gray,ultra thick] (7.0,6.0) rectangle (49.0,48.0);
\foreach \x in {7,...,48}{
  \draw[gray,thin] (\x,54-\x) rectangle (\x+1,54-\x+1.0);
}
%%% layer 4
\draw[green,ultra thick,fill=white] (5.25,4.25) rectangle (47.25,46.25);
\foreach \x in {6,...,47}{
  \foreach \y in {4,...,45}{
    \draw[green,thin] (\x+0.25,\y+1.25) rectangle (\x-0.75,\y+0.25);
  }
}
%%% layer 3
\draw[brown,ultra thick,fill=white] (3.5,2.5) rectangle (45.5,44.5);
\foreach \x in {6,...,47}{
  \foreach \y in {4,...,45}{
    \draw[brown,thin] (\x-1.5,\y-0.5) rectangle (\x-2.5,\y-1.5);
  }
}
%%% layer 2
\draw[red,ultra thick,fill=white] (1.75,0.75) rectangle (43.75,42.75);
\foreach \x in {6,...,47}{
  \foreach \y in {4,...,45}{
    \draw[red,thin,opacity=0.5] (\x-3.25,\y-2.25) rectangle (\x-4.25,\y-3.25);
  }
}
%%% layer 1
\draw[blue,ultra thick,fill=white] (0.0,-1.0) rectangle (42.0,41.0);
\draw[step=1.0,blue,thin] (0.0,-1.0) grid (42.0,41.0);
% cutouts
\draw[black,ultra thick,fill=black] (10.0,10.0) rectangle (14.0,17.0);
%\node[color=white,rotate=90] at (12.0,13.5) {\large\tt cutout};
\draw[black,ultra thick,fill=black] (18.0,6.0) rectangle (22.0,10.0);
\draw[black,ultra thick,fill=black] (30.0,25.0) rectangle (36.0,29.0);
\node[color=white] at (33.0,27.0) {\large\tt cutout};
%%% row headings
\node at (-0.75,40.5) {\tt a};
\node at (-0.75,39.5) {\tt b};
\node at (-0.75,38.5) {\tt c};
\node at (-0.75,37.5) {\tt d};
\node at (-0.75,36.5) {\tt e};
\node at (-0.75,35.5) {\tt f};
\node at (-0.75,34.5) {\tt g};
\node at (-0.75,33.5) {\tt h};
\node at (-0.75,32.5) {\tt i};
\node at (-0.75,31.5) {\tt j};
\node at (-0.75,30.5) {\tt k};
\node at (-0.75,29.5) {\tt l};
\node at (-0.75,28.5) {\tt m};
\node at (-0.75,27.5) {\tt n};
\node at (-0.75,26.5) {\tt o};
\node at (-0.75,25.5) {\tt p};
\node at (-0.75,24.5) {\tt q};
\node at (-0.75,23.5) {\tt r};
\node at (-0.75,22.5) {\tt s};
\node at (-0.75,21.5) {\tt t};
\node at (-0.75,20.5) {\tt u};
\node at (-0.75,19.5) {\tt v};
\node at (-0.75,18.5) {\tt w};
\node at (-0.75,17.5) {\tt x};
\node at (-0.75,16.5) {\tt y};
\node at (-0.75,15.5) {\tt z};
\node at (-0.75,14.5) {\tt 0};
\node at (-0.75,13.5) {\tt 1};
\node at (-0.75,12.5) {\tt 2};
\node at (-0.75,11.5) {\tt 3};
\node at (-0.75,10.5) {\tt 4};
\node at (-0.75,9.5) {\tt 5};
\node at (-0.75,8.5) {\tt 6};
\node at (-0.75,7.5) {\tt 7};
\node at (-0.75,6.5) {\tt 8};
\node at (-0.75,5.5) {\tt 9};
\node at (-1.2,4.5) {tab};
\node at (-1.0,3.5) {l-s};
\node at (-1.0,2.5) {ba};
\node at (-1.0,1.5) {r-s};
\node at (-1.2,0.5) {cap};
\node at (-1.0,-0.5) {sp};
\node at (7.5,48.5) {\smash{\tt a}};
\node at (8.5,48.5) {\smash{\tt b}};
\node at (9.5,48.5) {\smash{\tt c}};
\node at (10.5,48.5) {\smash{\tt d}};
\node at (11.5,48.5) {\smash{\tt e}};
\node at (12.5,48.5) {\smash{\tt f}};
\node at (13.5,48.5) {\smash{\tt g}};
\node at (14.5,48.5) {\smash{\tt h}};
\node at (15.5,48.5) {\smash{\tt i}};
\node at (16.5,48.5) {\smash{\tt j}};
\node at (17.5,48.5) {\smash{\tt k}};
\node at (18.5,48.5) {\smash{\tt l}};
\node at (19.5,48.5) {\smash{\tt m}};
\node at (20.5,48.5) {\smash{\tt n}};
\node at (21.5,48.5) {\smash{\tt o}};
\node at (22.5,48.5) {\smash{\tt p}};
\node at (23.5,48.5) {\smash{\tt q}};
\node at (24.5,48.5) {\smash{\tt r}};
\node at (25.5,48.5) {\smash{\tt s}};
\node at (26.5,48.5) {\smash{\tt t}};
\node at (27.5,48.5) {\smash{\tt u}};
\node at (28.5,48.5) {\smash{\tt v}};
\node at (29.5,48.5) {\smash{\tt w}};
\node at (30.5,48.5) {\smash{\tt x}};
\node at (31.5,48.5) {\smash{\tt y}};
\node at (32.5,48.5) {\smash{\tt z}};
\node at (33.5,48.5) {\smash{\tt 0}};
\node at (34.5,48.5) {\smash{\tt 1}};
\node at (35.5,48.5) {\smash{\tt 2}};
\node at (36.5,48.5) {\smash{\tt 3}};
\node at (37.5,48.5) {\smash{\tt 4}};
\node at (38.5,48.5) {\smash{\tt 5}};
\node at (39.5,48.5) {\smash{\tt 6}};
\node at (40.5,48.5) {\smash{\tt 7}};
\node at (41.5,48.5) {\smash{\tt 8}};
\node at (42.5,48.5) {\smash{\tt 9}};
\node[rotate=90] at (43.5,49.25) {tab};
\node[rotate=90] at (44.5,49.05) {l-s};
\node[rotate=90] at (45.5,49.05) {ba};
\node[rotate=90] at (46.5,49.05) {r-s};
\node[rotate=90] at (47.5,49.3) {cap};
\node[rotate=90] at (48.5,49.035) {sp};
\end{tikzpicture}
\caption{Keystroke dynamics image for free-text} 
\label{fig:keystorke_dynamics_image}
\end{figure}
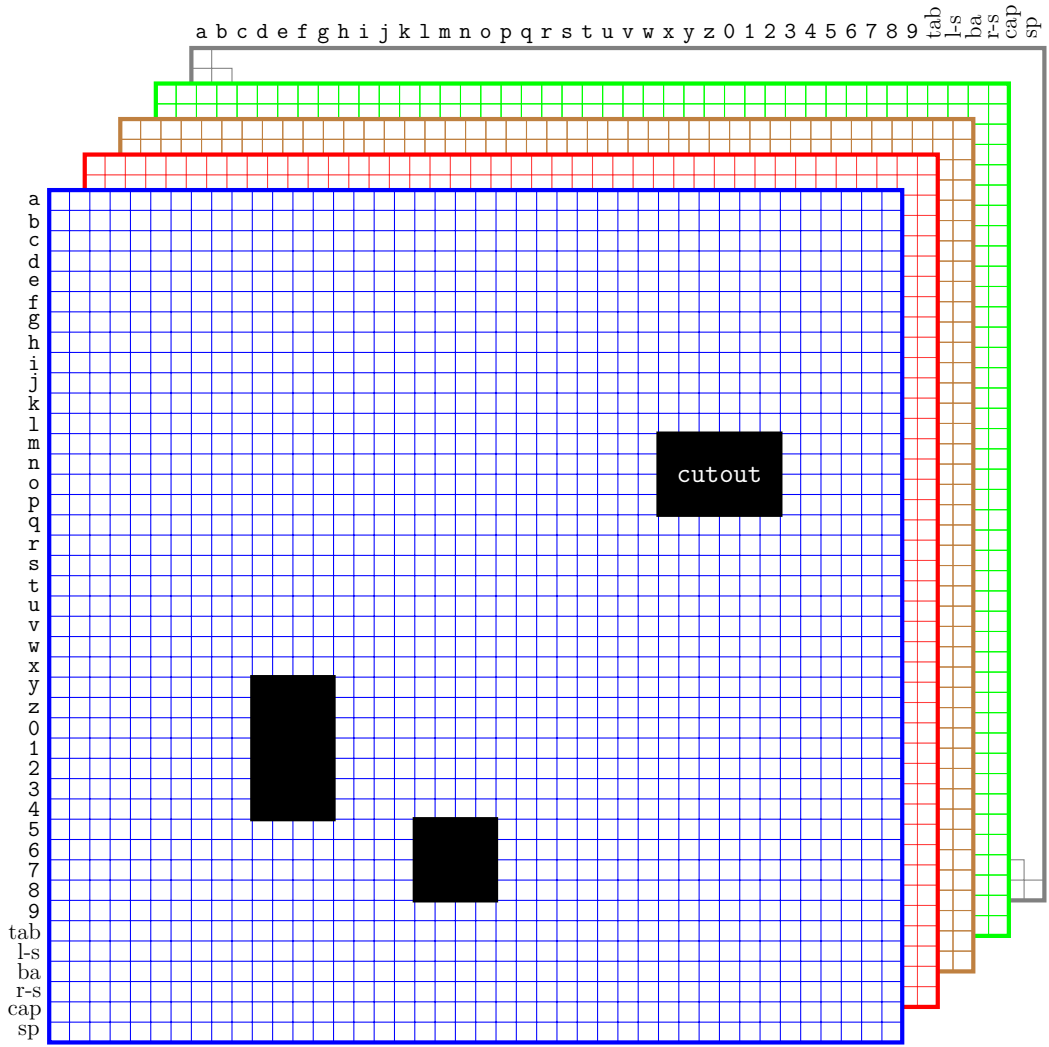

Each row and each column in our four-channel~$N\times N$ feature matrix corresponds 
to a key on the keyboard, and each channel corresponds to one kind of feature. 
Figure~\ref{fig:keystorke_dynamics_image} illustrates how we have organized 
these features into transition matrices. For example, the value at row~\texttt{i} and 
column~\texttt{j} in the first channel of the matrix refer to the UD-time between any
key presses of~\texttt{i} followed by~\texttt{j} within the current observation window. 

The final feature is duration, which is organized as a diagonal matrix and added to 
the transition matrix as a fifth channel. Note that if a key or key-pair is pressed more than
once, we use the average as the duration for that key or key-pair.
In this channel, only diagonal locations have 
values because the duration feature is only relevant for one key at a time. The final
result is that all of the features generated from keystroke subsequence are embedded in 
a transition matrix with five channels, which we refer to as the keystroke dynamics image (KDI).

To prevent the transition matrix from being too sparse, we only consider time-based features 
for the~42 most common keystrokes. These~42 keys include the~26 English characters (A-Z), 
the~10 Arabic numerals (0-9), and six meta keys (space, back, left-shift, right-shift, tab, and capital). 
Therefore, the shape of the transition matrix is~$5\times 42\times 42$,
with the five channels as discussed above. 

\subsection{Keystroke Dynamics Sequence}\label{sec:ksequence}

Above, we provided details on the time-based image-like feature that we construct, 
which we refer to as the KDI.
In this section, we discuss the application of an RNN-based neural network to the KDI.
Our goal is to use this feature to better take advantage of the inherently sequential nature 
of the keystroke dynamics data.

A keystroke in a keystroke sequence can be viewed as a word in a sentence. 
For our two free-text keystroke datasets, the keystroke sequence is different for each input 
and each user. For this data, we consider various encodings of each keystroke and 
use this encoding information in the embedding vector. Specifically, we experiment with 
index encoding and one-hot encoding.
The resulting embedding vectors are used to construct a keystroke dynamics sequence,
which we abbreviate as KDS. These KDS vectors will be used in 
our RNN-based neural networks.

\subsection{Cutout Regularization}\label{sec:augmentation}

As mentioned in Section~\ref{sec:cutout}, we employ a cutout regularization
to prevent overfitting in our CNN. By artificially adding occlusions to our image-like data, 
the network is forced to pay attention to all parts of the image, instead of over-emphasizing 
some specific parts. We apply cutouts to our novel KDI data structure,
which is discussed in Section~\ref{sec:kimage}, and the KDS, 
which was mentioned in Section~\ref{sec:ksequence}. 
The dark blocks in Figure~\ref{fig:keystorke_dynamics_image} 
illustrate cutouts.

\section{Architecture}\label{chap:architeture}

In this section, we discuss the classification 
models in more detail. 
We also discuss hyperparameters tuning for the models considered.

\subsection{Multiclass vs Binary Classification}

The Buffalo and Clarkson~II keystroke datasets are based on~101, and~148 subjects, 
respectively. Regardless of the dataset, our goal is to verify a user's identity based on features 
derived from keystroke sequences. While this is a classification problem, 
we can consider it as either a multiclass problem or multiple binary classification problems. 
In a practical application, the number of users could be orders of magnitude higher than
in either of our datasets. To train a multiclass model on a large number of users 
would be extremely costly, and each time a new user joins, the entire model would
have to be retrained. This is clearly impractical.

To train and test our models, we require positive and negative samples for each user. 
All the data available for a specific user will be considered as positive samples, while an equivalent 
number of negative samples are selected at random (and proportionally)
from other users' samples. In practice, the
number of non-target users may be very large. In that case, we could draw negative samples
from a a fixed number of non-target users.

\subsection{Hyperparameter Tuning}\label{sec:hyperparameters-tunning}

For the deep learning methods used in our experiments,
we employ a grid search to find the best parameters for 
the initial learning rate, optimizer, number of epochs, 
and learning rate schedule. The values shown in Table~\ref{tab:3.0} 
were tested, and those in boldface were 
found to generate the best result. To allow for a direct comparison of our different
models, we use these same hyperparameters for all of our deep learning models. 
Note that a learning rate of~0.01 generates the best results for CNN, MLP, LSTM, GRU,
while a learning rate of~0.001 generates best result in our RNN experiments.

\begin{table}[!htb]
\caption{Best Hyperparamters of deep learning models}\label{tab:3.0}
\centering
\adjustbox{scale=0.85}{
\begin{tabular}{c|c}\midrule\midrule
Parameter & Search space\\ \midrule
Training epochs & 100, \textbf{200}, 500, 1000 \\
Initial learning rate & 0.1, \textbf{0.01}, \textbf{0.001}, 0.0001 \\
Optimizer & \textbf{Adam}, SGD, SGD with Momentum \\
Learning schedule & \textbf{StepLR} (\textbf{0.1}, 0.3, 0.5), Plateau \\  
Experiments & \textbf{50} \\ \midrule\midrule
\end{tabular}
}
\end{table}

\subsection{Implementations}

For our keystroke dynamics experiments, we evaluate two kinds of models. 
Specifically, a CNN is applied to the our novel KDI image-like features,
while a hybrid model that combines CNN and GRU is applied to the KDS features.
The KDI is presented in Section~\ref{sec:kimage}, while the free-KDS is described in
Section~\ref{sec:ksequence}.

\subsubsection{CNN}

\begin{figure}[!htb]
\centering
\includegraphics[width=0.75\textwidth]{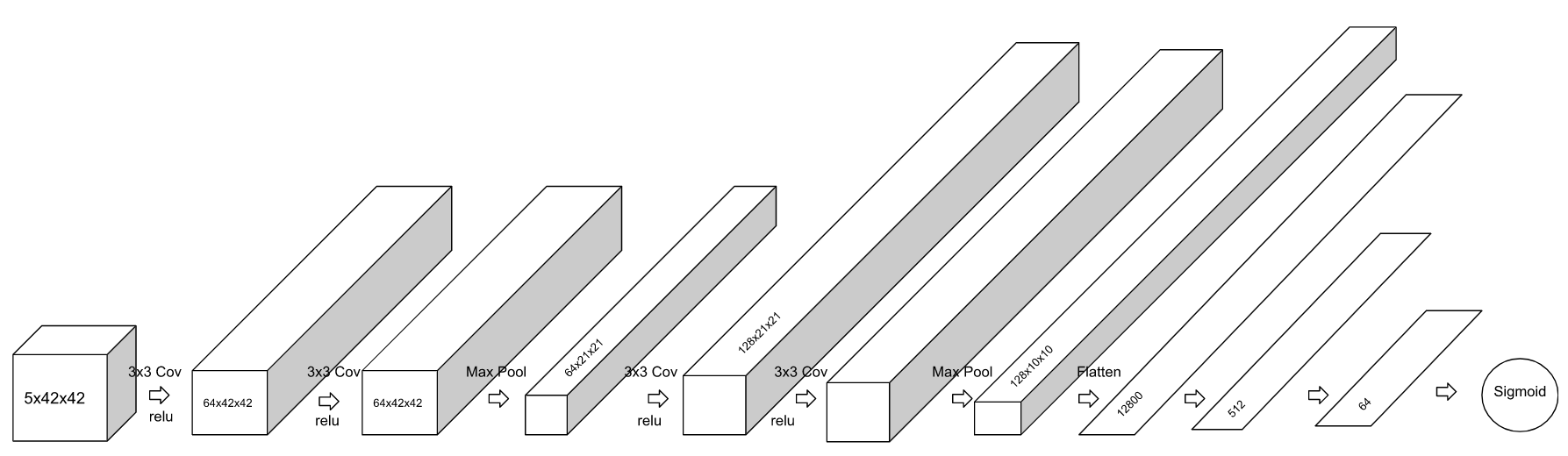}
\caption{Architecture of CNN for free-text datasets}\label{fig:cnn}
\end{figure}

The architecture of our CNN is shown in Figure~\ref{fig:cnn}. 
The input of this model is the KDI, and hence we view the transition matrix as an image. 
Here, a ``stage'' includes two \texttt{conv2d} layers and a \texttt{maxpooling} layer,
not counting the activation function.

In each stage, there are two convolutional layers and a maxpooling layer. 
Moreover, a \texttt{relu} function is employed after each convolutional layer.
Following these two stages, there are three fully connected layers, 
and a dropout layer is added to prevent overfitting.
Finally, a sigmoid function is used to compute the final probability of a positive sample. 

\subsubsection{CNN-RNN}

\begin{figure}[!htb]
\centering
\includegraphics[width=0.85\textwidth]{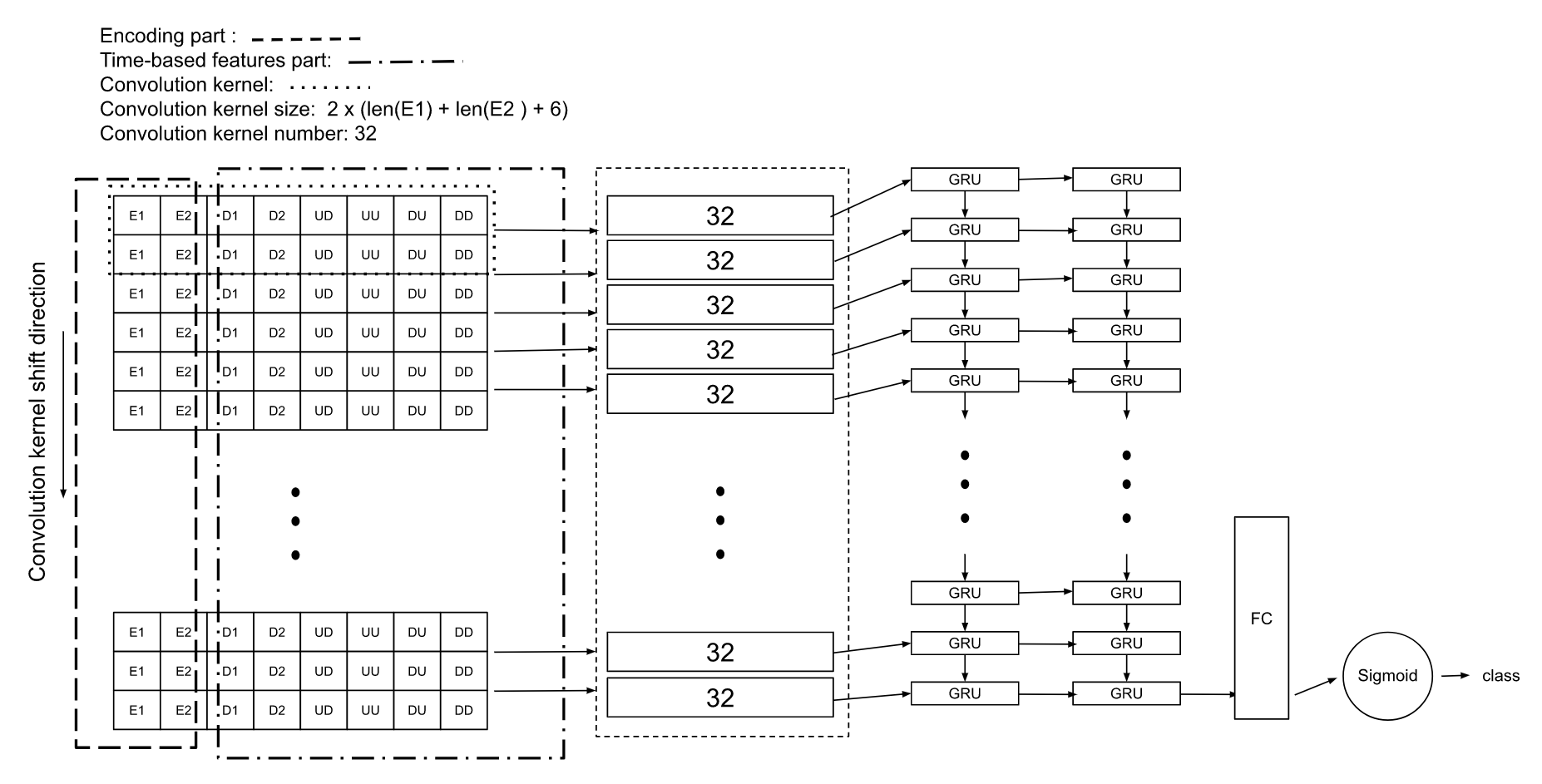}
\caption{Architecture of CNN-RNN for free-text datasets}\label{fig:cnn_rnn_architecture}
\end{figure}

The architecture of our CNN-RNN is illustrated in Figure~\ref{fig:cnn_rnn_architecture}. 
The input to this model is the KDS mentioned in Section~\ref{sec:ksequence}. 
Note that~32 convolutional kernels 
shift in the keystroke sequence direction, and
thus a sequence matrix with embedding size~32 is generated. 
This resulting output matrix is fed into a 2-layers GRU network,
which is followed by a fully connected layer.
Since this is a binary classification model, a sigmoid function is 
used to compute the probability of a positive sample.

\section{Experiment and Result}\label{chap:result}

In this section, we provide experimental results for 
our free-text binary classification experiments.
The results of the various models considered are analyzed and compared.
Note that in all of our experiments, 
we apply 5-folds cross validation
and average the performance for each user. 

\subsection{Metrics}

We adopt two metrics to evaluate our results. The first metric is accuracy, 
which is simply the number of correct classifications divided by the total
number of classifications. More formally, accuracy is calculated as
$$
  \mbox{accuracy}= \frac{\TP+\TN}{\TP+\FP+\TN+\FN}
$$
where~TP and~TN are true positives and true negatives, while~FP and~FN 
are false positives and false negatives.

\begin{figure}[!htb]
\centering
\includegraphics[width=0.5\textwidth]{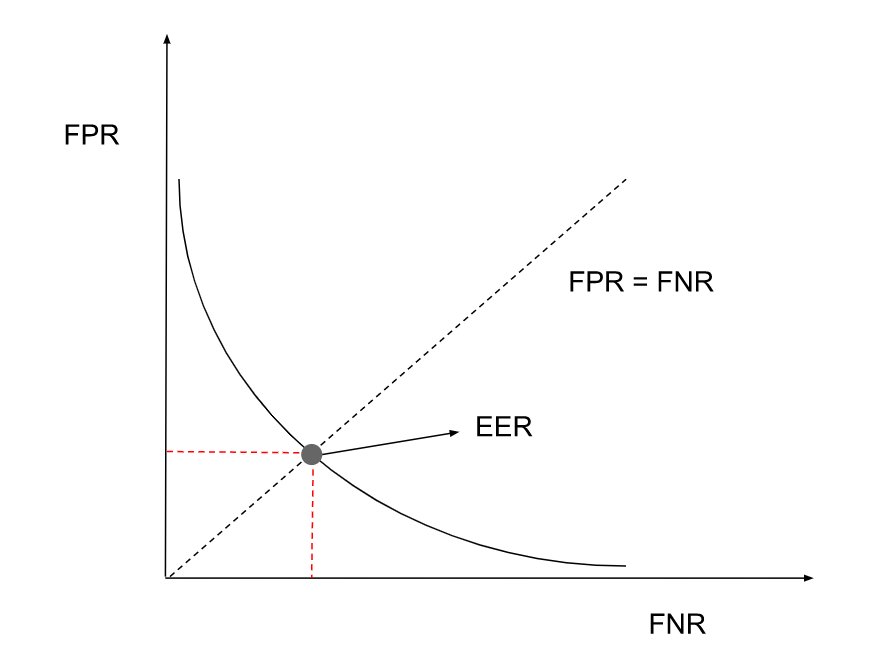}
\caption{Equal error rate} 
\label{fig:eer}
\end{figure}

There are two kinds of classification errors, namely, false positives and false negatives.
There is an inherent trade-off between the false positive rate (FPR) and the false negative rate (FNR),
in the sense that by changing the threshold that we use
for classification, we can lower one but the other will rise. 
For a metric that is threshold-independent, we compute the
equal error rate (EER) which, as the name suggests, is the value for which the FPR and FNR are equal.
The EER is obtained by considering thresholds in the range of~$[0, 1]$ to find the
point where the FPR and FNR are in balance. Figure~\ref{fig:eer} illustrates 
a technique for determining the EER.

\subsection{Result of Free-Text Experiments}\label{sec:results-free}

For our free-text experiments, we focus on the effect of the lengths of keystroke sequences,
kernel sizes for the CNN,  encoding methods for the keystroke sequence data, 
and different hyperparameters of the RNN. Additionally, we explore the performance of models 
with and without cutout regularization. 
%We also compare our results to previous work.

\subsubsection{Length of Keystroke Subsequence}

\begin{figure}[!htb]
\centering
\begin{tabular}{cc}
\begin{tikzpicture}[scale=0.55, every node/.style={scale=1.0}]
\begin{axis}[
xbar,
axis x line*=bottom,
width  = 0.65*\textwidth,
height = 7.5cm,
xmin=0,xmax=0.32,
xtick={0.00,0.05,0.10,0.15,0.20,0.25,0.30},
xlabel = {EER},
major y tick style = transparent,
xbar=5*\pgflinewidth,
bar width=9.0pt,
bar shift=-5.35pt,
x tick label style={
    	/pgf/number format/.cd,
   	fixed,
   	fixed zerofill,
    	precision=2},
y tick label style={
		font=\footnotesize,
%		anchor=north east,
%		inner sep=0mm
		},
symbolic y coords={%
	CNN length 50,
	CNN length 75,
	CNN length 100,
	CNN-RNN length 50,
	CNN-RNN length 75,
	CNN-RNN length 100
	},
ytick=data,
enlarge y limits=0.125,
%legend cell align=left,
%legend style={
%%	at={(1,1.05)},
%%	anchor=south east,
%%	nodes={rotate=90},%%%%% rotate text in legend
%%	at={(0.125,0)},
%%	at={(0.125,0)},
%%	at={(0.8775,0)},
%	at={(0.225,0.0175)},
%	anchor=south,
%	column sep=1ex
%},
nodes near coords, 
%nodes near coords align={horizontal},
every node near coord/.append style={
	anchor=west,
	font=\footnotesize,
	/pgf/number format/.cd,
	fixed,
	fixed zerofill,
	precision=4},
ytick=data,
]
\addplot [fill=red,opacity=1.00]
coordinates {
(0.0088,CNN length 50)
(0.0057,CNN length 75)
(0.0078,CNN length 100)
(0.0085,CNN-RNN length 50)
(0.0104,CNN-RNN length 75)
(0.0142,CNN-RNN length 100)
};
%\addlegendentry{EER}
\label{EERplotLB}
\end{axis}
%
%
%%%%%%%%%%%%%%%%%%%%%%%%%%%%%%%%%%%%%%%%%%
%
%
\begin{axis}[ 
xbar,
axis x line*=top,
width  = 0.65*\textwidth,
height = 7.5cm,
xmin=0,xmax=112.5,
xtick={0,20,40,60,80,100},
xlabel = {Accuracy (percentage)},
major y tick style = transparent,
xbar=5*\pgflinewidth,
bar width=9.0pt,
bar shift=5.35pt,
x tick label style={
    	/pgf/number format/.cd,
   	fixed,
   	fixed zerofill,
    	precision=0},
y tick label style={
		font=\footnotesize,
%		anchor=north east,
%		inner sep=0mm
		},
symbolic y coords={%
	CNN length 50,
	CNN length 75,
	CNN length 100,
	CNN-RNN length 50,
	CNN-RNN length 75,
	CNN-RNN length 100
	},
ytick=data,
enlarge y limits=0.125,
legend cell align=left,
legend style={
%	at={(1,1.05)},
%	anchor=south east,
%	nodes={rotate=90},%%%%% rotate text in legend
%	at={(0.125,0)},
%	at={(0.125,0)},
%	at={(0.8775,0)},
%	at={(0.825,0.0175)},
%	at={(0.155,0.825)},
%	at={(0.5,0.825)},
	at={(0.5,0.79)},
	anchor=south,
	column sep=1ex
},
nodes near coords, 
%nodes near coords align={horizontal},
every node near coord/.append style={
	anchor=west,
	font=\footnotesize,
	/pgf/number format/.cd,
	fixed,
	fixed zerofill,
	precision=2},
ytick=data,
]
%\addplot [fill=blue,opacity=0.10]
%coordinates {
%(99.06,CNN-GRU-without-sampler-fine-tune)
%(98.24,CNN-GRU-with-sampler)
%(98.20,CNN-GRU-without-sampler-not-best-val)
%(97.84,CNN-GRU-without-sampler-best-EER)
%(98.25,CNN-GRU-without-sampler-best-val)
%(98.35,CNN-GRU-cross-entropy-loss)
%(89.89,CNN-GRU-transformer-encoder)
%(94.39,CNN-GRU-pre-trained-word-embedding)
%};
%\end{axis}
%\addlegendentry{EER}
\addplot [fill=blue,opacity=1.00]
coordinates {
(98.56,CNN length 50)
(98.75,CNN length 75)
(98.53,CNN length 100)
(98.50,CNN-RNN length 50)
(98.21,CNN-RNN length 75)
(97.68,CNN-RNN length 100)
};
\addlegendentry{Accuracy}
\addlegendimage{/pgfplots/refstyle=EERplotLB,color=black,fill=red}\addlegendentry{EER}
\end{axis}
\end{tikzpicture}
&
\begin{tikzpicture}[scale=0.55, every node/.style={scale=1.0}]
\begin{axis}[
xbar,
axis x line*=bottom,
width  = 0.65*\textwidth,
height = 7.5cm,
xmin=0,xmax=0.32,
xtick={0.00,0.05,0.10,0.15,0.20,0.25,0.30},
xlabel = {EER},
major y tick style = transparent,
xbar=5*\pgflinewidth,
bar width=9.0pt,
bar shift=-5.35pt,
x tick label style={
    	/pgf/number format/.cd,
   	fixed,
   	fixed zerofill,
    	precision=2},
y tick label style={
		font=\footnotesize,
%		anchor=north east,
%		inner sep=0mm
		},
symbolic y coords={%
	CNN length 50,
	CNN length 75,
	CNN length 100,
	CNN-RNN length 50,
	CNN-RNN length 75,
	CNN-RNN length 100
	},
ytick=data,
enlarge y limits=0.125,
%legend cell align=left,
%legend style={
%%	at={(1,1.05)},
%%	anchor=south east,
%%	nodes={rotate=90},%%%%% rotate text in legend
%%	at={(0.125,0)},
%%	at={(0.125,0)},
%%	at={(0.8775,0)},
%	at={(0.225,0.0175)},
%	anchor=south,
%	column sep=1ex
%},
nodes near coords, 
%nodes near coords align={horizontal},
every node near coord/.append style={
	anchor=west,
	font=\footnotesize,
	/pgf/number format/.cd,
	fixed,
	fixed zerofill,
	precision=4},
ytick=data,
]
\addplot [fill=red,opacity=1.00]
coordinates {
(0.0743,CNN length 50)
(0.0731,CNN length 75)
(0.0715,CNN length 100)
(0.0774,CNN-RNN length 50)
(0.0894,CNN-RNN length 75)
(0.1061,CNN-RNN length 100)
};
%\addlegendentry{EER}
\label{EERplotLC}
\end{axis}
%
%
%%%%%%%%%%%%%%%%%%%%%%%%%%%%%%%%%%%%%%%%%%
%
%
\begin{axis}[ 
xbar,
axis x line*=top,
width  = 0.65*\textwidth,
height = 7.5cm,
xmin=0,xmax=112.5,
xtick={0,20,40,60,80,100},
xlabel = {Accuracy (percentage)},
major y tick style = transparent,
xbar=5*\pgflinewidth,
bar width=9.0pt,
bar shift=5.35pt,
x tick label style={
    	/pgf/number format/.cd,
   	fixed,
   	fixed zerofill,
    	precision=0},
y tick label style={
		font=\footnotesize,
%		anchor=north east,
%		inner sep=0mm
		},
symbolic y coords={%
	CNN length 50,
	CNN length 75,
	CNN length 100,
	CNN-RNN length 50,
	CNN-RNN length 75,
	CNN-RNN length 100
	},
ytick=data,
enlarge y limits=0.125,
legend cell align=left,
legend style={
%	at={(1,1.05)},
%	anchor=south east,
%	nodes={rotate=90},%%%%% rotate text in legend
%	at={(0.125,0)},
%	at={(0.125,0)},
%	at={(0.8775,0)},
%	at={(0.825,0.0175)},
%	at={(0.155,0.825)},
	at={(0.155,0.79)},
	anchor=south,
	column sep=1ex
},
nodes near coords, 
%nodes near coords align={horizontal},
every node near coord/.append style={
	anchor=west,
	font=\footnotesize,
	/pgf/number format/.cd,
	fixed,
	fixed zerofill,
	precision=2},
ytick=data,
]
%\addplot [fill=blue,opacity=0.10]
%coordinates {
%(99.06,CNN-GRU-without-sampler-fine-tune)
%(98.24,CNN-GRU-with-sampler)
%(98.20,CNN-GRU-without-sampler-not-best-val)
%(97.84,CNN-GRU-without-sampler-best-EER)
%(98.25,CNN-GRU-without-sampler-best-val)
%(98.35,CNN-GRU-cross-entropy-loss)
%(89.89,CNN-GRU-transformer-encoder)
%(94.39,CNN-GRU-pre-trained-word-embedding)
%};
%\end{axis}
%\addlegendentry{EER}
\addplot [fill=blue,opacity=1.00]
coordinates {
(91.82,CNN length 50)
(92.00,CNN length 75)
(92.09,CNN length 100)
(91.91,CNN-RNN length 50)
(90.53,CNN-RNN length 75)
(88.62,CNN-RNN length 100)
};
\addlegendentry{Accuracy}
\addlegendimage{/pgfplots/refstyle=EERplotLC,color=black,fill=red}\addlegendentry{EER}
\end{axis}
\end{tikzpicture}
\\
(a) Buffalo dataset
&
(b) Clarkson~II dataset
\end{tabular}
\caption{Keystroke lengths}\label{fig:5.2-3}
\end{figure}
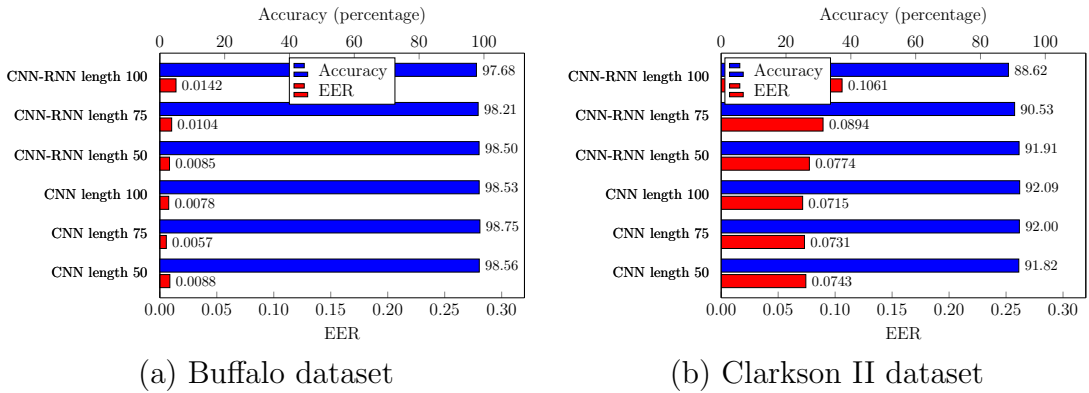

First, we experiment with different lengths of keystroke subsequences.
Specifically, we consider lengths of~50, 75, and~100 keystrokes.
The results of these experiment are given in Figures~\ref{fig:5.2-3}~(a) and~(b)
for the Buffalo and Clarkson~II datasets, respectively. 
From these results, we observe that when the length of a keystroke sequence 
has minimal impact on the accuracy or EER.

From these results, we observe that when the length of a keystroke sequence 
is relatively short, there is insufficient information to support strong authentication 
and, conversely, when the sequence is too long, the additional noise degrades 
the accuracy. Moreover, the results shows that the CNN-based model is more 
robust when the length of the keystroke sequence changes, which can be 
explained by the KDI mitigating the noise inherent in longer sequences.
To accelerate the training process, we adopt the length~100 for the keystroke 
subsequences in all subsequent experiments.

\subsubsection{CNN Kernel Sizes}

\begin{figure}[!htb]
\centering
\begin{tabular}{cc}
\begin{tikzpicture}[scale=0.55, every node/.style={scale=1.0}]
\begin{axis}[
xbar,
axis x line*=bottom,
width  = 0.65*\textwidth,
height = 7.5cm,
xmin=0,xmax=0.32,
xtick={0.00,0.05,0.10,0.15,0.20,0.25,0.30},
xlabel = {EER},
major y tick style = transparent,
xbar=5*\pgflinewidth,
bar width=9.0pt,
bar shift=-5.35pt,
x tick label style={
    	/pgf/number format/.cd,
   	fixed,
   	fixed zerofill,
    	precision=2},
y tick label style={
		font=\footnotesize,
%		anchor=north east,
%		inner sep=0mm
		},
symbolic y coords={%
	CNN $3\times 3$,
	CNN $5\times 5$,
	CNN $7\times 7$,
	CNN-RNN $2\times 8$,
	CNN-RNN $3\times 8$,
	CNN-RNN $5\times 8$
	},
ytick=data,
enlarge y limits=0.125,
%legend cell align=left,
%legend style={
%%	at={(1,1.05)},
%%	anchor=south east,
%%	nodes={rotate=90},%%%%% rotate text in legend
%%	at={(0.125,0)},
%%	at={(0.125,0)},
%%	at={(0.8775,0)},
%	at={(0.225,0.0175)},
%	anchor=south,
%	column sep=1ex
%},
nodes near coords, 
%nodes near coords align={horizontal},
every node near coord/.append style={
	anchor=west,
	font=\footnotesize,
	/pgf/number format/.cd,
	fixed,
	fixed zerofill,
	precision=4},
ytick=data,
]
\addplot [fill=red,opacity=1.00]
coordinates {
(0.0078,CNN $3\times 3$)
(0.0049,CNN $5\times 5$)
(0.0050,CNN $7\times 7$)
(0.0142,CNN-RNN $2\times 8$)
(0.0160,CNN-RNN $3\times 8$)
(0.0190,CNN-RNN $5\times 8$)
};
%\addlegendentry{EER}
\label{EERplotKB}
\end{axis}
%
%
%%%%%%%%%%%%%%%%%%%%%%%%%%%%%%%%%%%%%%%%%%
%
%
\begin{axis}[ 
xbar,
axis x line*=top,
width  = 0.65*\textwidth,
height = 7.5cm,
xmin=0,xmax=112.5,
xtick={0,20,40,60,80,100},
xlabel = {Accuracy (percentage)},
major y tick style = transparent,
xbar=5*\pgflinewidth,
bar width=9.0pt,
bar shift=5.35pt,
x tick label style={
    	/pgf/number format/.cd,
   	fixed,
   	fixed zerofill,
    	precision=0},
y tick label style={
		font=\footnotesize,
%		anchor=north east,
%		inner sep=0mm
		},
symbolic y coords={%
	CNN $3\times 3$,
	CNN $5\times 5$,
	CNN $7\times 7$,
	CNN-RNN $2\times 8$,
	CNN-RNN $3\times 8$,
	CNN-RNN $5\times 8$
	},
ytick=data,
enlarge y limits=0.125,
legend cell align=left,
legend style={
%	at={(1,1.05)},
%	anchor=south east,
%	nodes={rotate=90},%%%%% rotate text in legend
%	at={(0.125,0)},
%	at={(0.125,0)},
%	at={(0.8775,0)},
%	at={(0.825,0.0175)},
%	at={(0.155,0.825)},
%	at={(0.5,0.825)},
	at={(0.5,0.79)},
	anchor=south,
	column sep=1ex
},
nodes near coords, 
%nodes near coords align={horizontal},
every node near coord/.append style={
	anchor=west,
	font=\footnotesize,
	/pgf/number format/.cd,
	fixed,
	fixed zerofill,
	precision=2},
ytick=data,
]
%\addplot [fill=blue,opacity=0.10]
%coordinates {
%(99.06,CNN-GRU-without-sampler-fine-tune)
%(98.24,CNN-GRU-with-sampler)
%(98.20,CNN-GRU-without-sampler-not-best-val)
%(97.84,CNN-GRU-without-sampler-best-EER)
%(98.25,CNN-GRU-without-sampler-best-val)
%(98.35,CNN-GRU-cross-entropy-loss)
%(89.89,CNN-GRU-transformer-encoder)
%(94.39,CNN-GRU-pre-trained-word-embedding)
%};
%\end{axis}
%\addlegendentry{EER}
\addplot [fill=blue,opacity=1.00]
coordinates {
(98.53,CNN $3\times 3$)
(98.61,CNN $5\times 5$)
(98.51,CNN $7\times 7$)
(97.68,CNN-RNN $2\times 8$)
(97.53,CNN-RNN $3\times 8$)
(97.02,CNN-RNN $5\times 8$)
};
\addlegendentry{Accuracy}
\addlegendimage{/pgfplots/refstyle=EERplotKB,color=black,fill=red}\addlegendentry{EER}
\end{axis}
\end{tikzpicture}
&
\begin{tikzpicture}[scale=0.55, every node/.style={scale=1.0}]
\begin{axis}[
xbar,
axis x line*=bottom,
width  = 0.65*\textwidth,
height = 7.5cm,
xmin=0,xmax=0.32,
xtick={0.00,0.05,0.10,0.15,0.20,0.25,0.30},
xlabel = {EER},
major y tick style = transparent,
xbar=5*\pgflinewidth,
bar width=9.0pt,
bar shift=-5.35pt,
x tick label style={
    	/pgf/number format/.cd,
   	fixed,
   	fixed zerofill,
    	precision=2},
y tick label style={
		font=\footnotesize,
%		anchor=north east,
%		inner sep=0mm
		},
symbolic y coords={%
	CNN $3\times 3$,
	CNN $5\times 5$,
	CNN $7\times 7$,
	CNN-RNN $2\times 8$,
	CNN-RNN $3\times 8$,
	CNN-RNN $5\times 8$
	},
ytick=data,
enlarge y limits=0.125,
%legend cell align=left,
%legend style={
%%	at={(1,1.05)},
%%	anchor=south east,
%%	nodes={rotate=90},%%%%% rotate text in legend
%%	at={(0.125,0)},
%%	at={(0.125,0)},
%%	at={(0.8775,0)},
%	at={(0.225,0.0175)},
%	anchor=south,
%	column sep=1ex
%},
nodes near coords, 
%nodes near coords align={horizontal},
every node near coord/.append style={
	anchor=west,
	font=\footnotesize,
	/pgf/number format/.cd,
	fixed,
	fixed zerofill,
	precision=4},
ytick=data,
]
\addplot [fill=red,opacity=1.00]
coordinates {
(0.0716,CNN $3\times 3$)
(0.0904,CNN $5\times 5$)
(0.0990,CNN $7\times 7$)
(0.1061,CNN-RNN $2\times 8$)
(0.1277,CNN-RNN $3\times 8$)
(0.1650,CNN-RNN $5\times 8$)
};
%\addlegendentry{EER}
\label{EERplotKC}
\end{axis}
%
%
%%%%%%%%%%%%%%%%%%%%%%%%%%%%%%%%%%%%%%%%%%
%
%
\begin{axis}[ 
xbar,
axis x line*=top,
width  = 0.65*\textwidth,
height = 7.5cm,
xmin=0,xmax=112.5,
xtick={0,20,40,60,80,100},
xlabel = {Accuracy (percentage)},
major y tick style = transparent,
xbar=5*\pgflinewidth,
bar width=9.0pt,
bar shift=5.35pt,
x tick label style={
    	/pgf/number format/.cd,
   	fixed,
   	fixed zerofill,
    	precision=0},
y tick label style={
		font=\footnotesize,
%		anchor=north east,
%		inner sep=0mm
		},
symbolic y coords={%
	CNN $3\times 3$,
	CNN $5\times 5$,
	CNN $7\times 7$,
	CNN-RNN $2\times 8$,
	CNN-RNN $3\times 8$,
	CNN-RNN $5\times 8$
	},
ytick=data,
enlarge y limits=0.125,
legend cell align=left,
legend style={
%	at={(1,1.05)},
%	anchor=south east,
%	nodes={rotate=90},%%%%% rotate text in legend
%	at={(0.125,0)},
%	at={(0.125,0)},
%	at={(0.8775,0)},
%	at={(0.825,0.0175)},
%	at={(0.155,0.825)},
%	at={(0.5,0.825)},
	at={(0.5,0.79)},
	anchor=south,
	column sep=1ex
},
nodes near coords, 
%nodes near coords align={horizontal},
every node near coord/.append style={
	anchor=west,
	font=\footnotesize,
	/pgf/number format/.cd,
	fixed,
	fixed zerofill,
	precision=2},
ytick=data,
]
%\addplot [fill=blue,opacity=0.10]
%coordinates {
%(99.06,CNN-GRU-without-sampler-fine-tune)
%(98.24,CNN-GRU-with-sampler)
%(98.20,CNN-GRU-without-sampler-not-best-val)
%(97.84,CNN-GRU-without-sampler-best-EER)
%(98.25,CNN-GRU-without-sampler-best-val)
%(98.35,CNN-GRU-cross-entropy-loss)
%(89.89,CNN-GRU-transformer-encoder)
%(94.39,CNN-GRU-pre-trained-word-embedding)
%};
%\end{axis}
%\addlegendentry{EER}
\addplot [fill=blue,opacity=1.00]
coordinates {
(92.09,CNN $3\times 3$)
(90.68,CNN $5\times 5$)
(84.79,CNN $7\times 7$)
(88.62,CNN-RNN $2\times 8$)
(86.61,CNN-RNN $3\times 8$)
(82.84,CNN-RNN $5\times 8$)
};
\addlegendentry{Accuracy}
\addlegendimage{/pgfplots/refstyle=EERplotKC,color=black,fill=red}\addlegendentry{EER}
\end{axis}
\end{tikzpicture}
\\
(a) Buffalo dataset
&
(b) Clarkson~II dataset
\end{tabular}
\caption{Kernel size}\label{fig:5.4-5}
\end{figure}
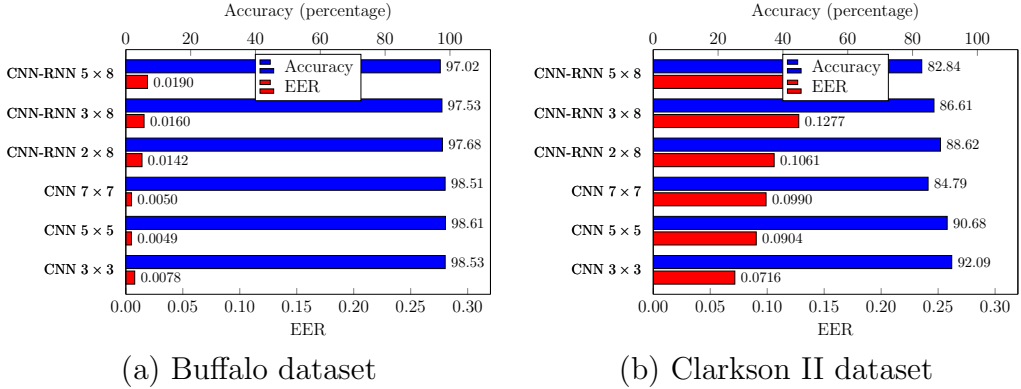

In any CNN, the kernel size is a critical parameter. To determine the optimal kernel size,
we experiment with three square kernels ($3 \times 3$, $5 \times 5$, and~$7 \times 7$) 
in our basic CNN model. For the CNN part of our hybrid CNN-RNN model,
we experiment with three sizes of rectangle kernels ($2 \times 8$, $3 \times 8$, and~$5 \times 8$). 
These experimental results for the basic CNN and hybrid CNN-RNN
are given in Figure~\ref{fig:5.4-5}.

We note that the kernel size makes no appreciable difference for the basic CNN model
on the Buffalo dataset, while the two larger kernels both perform equally well
on the Clarkson~II dataset. For the CNN-RNN model, the results are also mixed,
with the smaller kernel giving the best results over the two datasets.
We adopt~$3 \times 3$ square kernels for CNN-based models and~$2 \times 8$ 
kernels for CNN-RNN based model in subsequent experiments.

\subsubsection{Embedding Method}

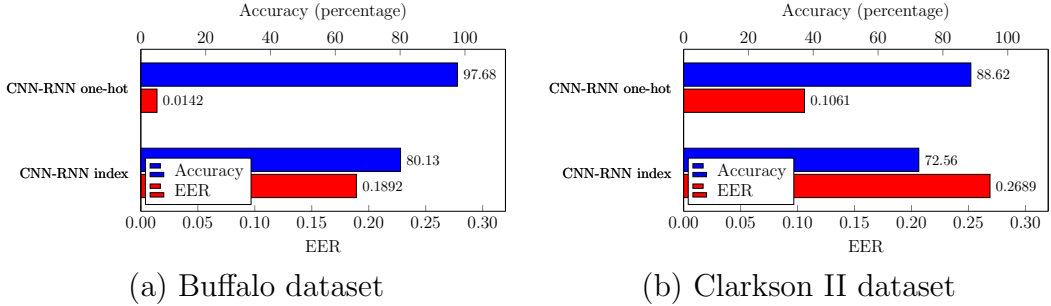
\begin{figure}[!htb]
\centering
\begin{tabular}{cc}
\begin{tikzpicture}[scale=0.55, every node/.style={scale=1.0}]
\begin{axis}[
xbar,
axis x line*=bottom,
width  = 0.65*\textwidth,
height = 5.5cm,
xmin=0,xmax=0.32,
xtick={0.00,0.05,0.10,0.15,0.20,0.25,0.30},
xlabel = {EER},
major y tick style = transparent,
xbar=5*\pgflinewidth,
bar width=16.0pt,
bar shift=-9.0pt,
x tick label style={
    	/pgf/number format/.cd,
   	fixed,
   	fixed zerofill,
    	precision=2},
y tick label style={
		font=\footnotesize,
%		anchor=north east,
%		inner sep=0mm
		},
symbolic y coords={%
	CNN-RNN index,
	CNN-RNN one-hot
	},
ytick=data,
enlarge y limits=0.45,
%legend cell align=left,
%legend style={
%%	at={(1,1.05)},
%%	anchor=south east,
%%	nodes={rotate=90},%%%%% rotate text in legend
%%	at={(0.125,0)},
%%	at={(0.125,0)},
%%	at={(0.8775,0)},
%	at={(0.225,0.0175)},
%	anchor=south,
%	column sep=1ex
%},
nodes near coords, 
%nodes near coords align={horizontal},
every node near coord/.append style={
	anchor=west,
	font=\footnotesize,
	/pgf/number format/.cd,
	fixed,
	fixed zerofill,
	precision=4},
ytick=data,
]
\addplot [fill=red,opacity=1.00]
coordinates {
(0.1892,CNN-RNN index)
(0.0142,CNN-RNN one-hot)
};
%\addlegendentry{EER}
\label{EERplotEB}
\end{axis}
%
%
%%%%%%%%%%%%%%%%%%%%%%%%%%%%%%%%%%%%%%%%%%
%
%
\begin{axis}[ 
xbar,
axis x line*=top,
width  = 0.65*\textwidth,
height = 5.5cm,
xmin=0,xmax=112.5,
xtick={0,20,40,60,80,100},
xlabel = {Accuracy (percentage)},
major y tick style = transparent,
xbar=5*\pgflinewidth,
bar width=16.0pt,
bar shift=9.0pt,
x tick label style={
    	/pgf/number format/.cd,
   	fixed,
   	fixed zerofill,
    	precision=0},
y tick label style={
		font=\footnotesize,
%		anchor=north east,
%		inner sep=0mm
		},
symbolic y coords={%
	CNN-RNN index,
	CNN-RNN one-hot
	},
ytick=data,
enlarge y limits=0.45,
legend cell align=left,
legend style={
%	at={(1,1.05)},
%	anchor=south east,
%	nodes={rotate=90},%%%%% rotate text in legend
%	at={(0.125,0)},
%	at={(0.125,0)},
%	at={(0.8775,0)},
%	at={(0.825,0.0175)},
%	at={(0.155,0.825)},
%	at={(0.5,0.825)},
%	at={(0.5,0.675)},
%	at={(0.165,0.05)},
	at={(0.1575,0.04)},
	anchor=south,
	column sep=1ex
},
nodes near coords, 
%nodes near coords align={horizontal},
every node near coord/.append style={
	anchor=west,
	font=\footnotesize,
	/pgf/number format/.cd,
	fixed,
	fixed zerofill,
	precision=2},
ytick=data,
]
%\addplot [fill=blue,opacity=0.10]
%coordinates {
%(99.06,CNN-GRU-without-sampler-fine-tune)
%(98.24,CNN-GRU-with-sampler)
%(98.20,CNN-GRU-without-sampler-not-best-val)
%(97.84,CNN-GRU-without-sampler-best-EER)
%(98.25,CNN-GRU-without-sampler-best-val)
%(98.35,CNN-GRU-cross-entropy-loss)
%(89.89,CNN-GRU-transformer-encoder)
%(94.39,CNN-GRU-pre-trained-word-embedding)
%};
%\end{axis}
%\addlegendentry{EER}
\addplot [fill=blue,opacity=1.00]
coordinates {
(80.13,CNN-RNN index)
(97.68,CNN-RNN one-hot)
};
\addlegendentry{Accuracy}
\addlegendimage{/pgfplots/refstyle=EERplotEB,color=black,fill=red}\addlegendentry{EER}
\end{axis}
\end{tikzpicture}
&
\begin{tikzpicture}[scale=0.55, every node/.style={scale=1.0}]
\begin{axis}[
xbar,
axis x line*=bottom,
width  = 0.65*\textwidth,
height = 5.5cm,
xmin=0,xmax=0.32,
xtick={0.00,0.05,0.10,0.15,0.20,0.25,0.30},
xlabel = {EER},
major y tick style = transparent,
xbar=5*\pgflinewidth,
bar width=16.0pt,
bar shift=-9.0pt,
x tick label style={
    	/pgf/number format/.cd,
   	fixed,
   	fixed zerofill,
    	precision=2},
y tick label style={
		font=\footnotesize,
%		anchor=north east,
%		inner sep=0mm
		},
symbolic y coords={%
	CNN-RNN index,
	CNN-RNN one-hot
	},
ytick=data,
enlarge y limits=0.45,
%legend cell align=left,
%legend style={
%%	at={(1,1.05)},
%%	anchor=south east,
%%	nodes={rotate=90},%%%%% rotate text in legend
%%	at={(0.125,0)},
%%	at={(0.125,0)},
%%	at={(0.8775,0)},
%	at={(0.225,0.0175)},
%	anchor=south,
%	column sep=1ex
%},
nodes near coords, 
%nodes near coords align={horizontal},
every node near coord/.append style={
	anchor=west,
	font=\footnotesize,
	/pgf/number format/.cd,
	fixed,
	fixed zerofill,
	precision=4},
ytick=data,
]
\addplot [fill=red,opacity=1.00]
coordinates {
(0.2689,CNN-RNN index)
(0.1061,CNN-RNN one-hot)
};
%\addlegendentry{EER}
\label{EERplotEC}
\end{axis}
%
%
%%%%%%%%%%%%%%%%%%%%%%%%%%%%%%%%%%%%%%%%%%
%
%
\begin{axis}[ 
xbar,
axis x line*=top,
width  = 0.65*\textwidth,
height = 5.5cm,
xmin=0,xmax=112.5,
xtick={0,20,40,60,80,100},
xlabel = {Accuracy (percentage)},
major y tick style = transparent,
xbar=5*\pgflinewidth,
bar width=16.0pt,
bar shift=9.0pt,
x tick label style={
    	/pgf/number format/.cd,
   	fixed,
   	fixed zerofill,
    	precision=0},
y tick label style={
		font=\footnotesize,
%		anchor=north east,
%		inner sep=0mm
		},
symbolic y coords={%
	CNN-RNN index,
	CNN-RNN one-hot
	},
ytick=data,
enlarge y limits=0.45,
legend cell align=left,
legend style={
%	at={(1,1.05)},
%	anchor=south east,
%	nodes={rotate=90},%%%%% rotate text in legend
%	at={(0.125,0)},
%	at={(0.125,0)},
%	at={(0.8775,0)},
%	at={(0.825,0.0175)},
%	at={(0.155,0.825)},
%	at={(0.5,0.825)},
%	at={(0.5,0.675)},
	at={(0.1575,0.04)},
	anchor=south,
	column sep=1ex
},
nodes near coords, 
%nodes near coords align={horizontal},
every node near coord/.append style={
	anchor=west,
	font=\footnotesize,
	/pgf/number format/.cd,
	fixed,
	fixed zerofill,
	precision=2},
ytick=data,
]
%\addplot [fill=blue,opacity=0.10]
%coordinates {
%(99.06,CNN-GRU-without-sampler-fine-tune)
%(98.24,CNN-GRU-with-sampler)
%(98.20,CNN-GRU-without-sampler-not-best-val)
%(97.84,CNN-GRU-without-sampler-best-EER)
%(98.25,CNN-GRU-without-sampler-best-val)
%(98.35,CNN-GRU-cross-entropy-loss)
%(89.89,CNN-GRU-transformer-encoder)
%(94.39,CNN-GRU-pre-trained-word-embedding)
%};
%\end{axis}
%\addlegendentry{EER}
\addplot [fill=blue,opacity=1.00]
coordinates {
(72.56,CNN-RNN index)
(88.62,CNN-RNN one-hot)
};
\addlegendentry{Accuracy}
\addlegendimage{/pgfplots/refstyle=EERplotEC,color=black,fill=red}\addlegendentry{EER}
\end{axis}
\end{tikzpicture}
\\
(a) Buffalo dataset
&
(b) Clarkson~II dataset
\end{tabular}
\caption{Embedding methods}\label{fig:5.6-7}
\end{figure}

As mentioned above in Section~\ref{sec:ksequence}, 
we consider two embedding methods, namely,
index encoding and one-hot encoding.
These experimental results are given in 
Figures~\ref{fig:5.6-7}~(a) and~(b) for the Buffalo and Clarkson~II datasets,
respectively. From these results, it is clear that one-hot encoding
is far superior to index encoding, and hence in subsequent experiments,
we use one-hot encoding. 

\subsubsection{RNN Structure}

We experiment with three types of RNN-based networks in our CNN-RNN architecture.
Specifically, we consider a plain RNN, GRU, and LSTM. 
The advantages of GRU and LSTM are that they can capture more
long-term information than a plain RNN. 
The results of these experiments are given in Figure~\ref{fig:5.8-9}.

\begin{figure}[!htb]
\centering
\begin{tabular}{cc}
\begin{tikzpicture}[scale=0.55, every node/.style={scale=1.0}]
\begin{axis}[
xbar,
axis x line*=bottom,
width  = 0.65*\textwidth,
height = 6.5cm,
xmin=0,xmax=0.32,
xtick={0.00,0.05,0.10,0.15,0.20,0.25,0.30},
xlabel = {EER},
major y tick style = transparent,
xbar=5*\pgflinewidth,
bar width=16.0pt,
bar shift=-9.0pt,
x tick label style={
    	/pgf/number format/.cd,
   	fixed,
   	fixed zerofill,
    	precision=2},
y tick label style={
		font=\footnotesize,
%		anchor=north east,
%		inner sep=0mm
		},
symbolic y coords={%
	CNN-RNN RNN,
	CNN-RNN LSTM,
	CNN-RNN GRU
	},
yticklabels={%
	CNN-RNN,
	CNN-LSTM,
	CNN-GRU
	},
ytick=data,
enlarge y limits=0.275,
%legend cell align=left,
%legend style={
%%	at={(1,1.05)},
%%	anchor=south east,
%%	nodes={rotate=90},%%%%% rotate text in legend
%%	at={(0.125,0)},
%%	at={(0.125,0)},
%%	at={(0.8775,0)},
%	at={(0.225,0.0175)},
%	anchor=south,
%	column sep=1ex
%},
nodes near coords, 
%nodes near coords align={horizontal},
every node near coord/.append style={
	anchor=west,
	font=\footnotesize,
	/pgf/number format/.cd,
	fixed,
	fixed zerofill,
	precision=4},
ytick=data,
]
\addplot [fill=red,opacity=1.00]
coordinates {
(0.0152,CNN-RNN RNN)
(0.0112,CNN-RNN LSTM)
(0.0142,CNN-RNN GRU)
};
%\addlegendentry{EER}
\label{EERplotRB}
\end{axis}
%
%
%%%%%%%%%%%%%%%%%%%%%%%%%%%%%%%%%%%%%%%%%%
%
%
\begin{axis}[ 
xbar,
axis x line*=top,
width  = 0.65*\textwidth,
height = 6.5cm,
xmin=0,xmax=112.5,
xtick={0,20,40,60,80,100},
xlabel = {Accuracy (percentage)},
major y tick style = transparent,
xbar=5*\pgflinewidth,
bar width=16.0pt,
bar shift=9.0pt,
x tick label style={
    	/pgf/number format/.cd,
   	fixed,
   	fixed zerofill,
    	precision=0},
y tick label style={
		font=\footnotesize,
%		anchor=north east,
%		inner sep=0mm
		},
symbolic y coords={%
	CNN-RNN RNN,
	CNN-RNN LSTM,
	CNN-RNN GRU
	},
yticklabels={%
	CNN-RNN,
	CNN-LSTM,
	CNN-GRU
	},
ytick=data,
enlarge y limits=0.275,
legend cell align=left,
legend style={
%	at={(1,1.05)},
%	anchor=south east,
%	nodes={rotate=90},%%%%% rotate text in legend
%	at={(0.125,0)},
%	at={(0.125,0)},
%	at={(0.8775,0)},
%	at={(0.825,0.0175)},
%	at={(0.155,0.825)},
%	at={(0.5,0.825)},
%	at={(0.5,0.675)},
%	at={(0.165,0.05)},
%	at={(0.1575,0.03)},
	at={(0.5,0.74)},
	anchor=south,
	column sep=1ex
},
nodes near coords, 
%nodes near coords align={horizontal},
every node near coord/.append style={
	anchor=west,
	font=\footnotesize,
	/pgf/number format/.cd,
	fixed,
	fixed zerofill,
	precision=2},
ytick=data,
]
%\addplot [fill=blue,opacity=0.10]
%coordinates {
%(99.06,CNN-GRU-without-sampler-fine-tune)
%(98.24,CNN-GRU-with-sampler)
%(98.20,CNN-GRU-without-sampler-not-best-val)
%(97.84,CNN-GRU-without-sampler-best-EER)
%(98.25,CNN-GRU-without-sampler-best-val)
%(98.35,CNN-GRU-cross-entropy-loss)
%(89.89,CNN-GRU-transformer-encoder)
%(94.39,CNN-GRU-pre-trained-word-embedding)
%};
%\end{axis}
%\addlegendentry{EER}
\addplot [fill=blue,opacity=1.00]
coordinates {
(97.46,CNN-RNN RNN)
(97.77,CNN-RNN LSTM)
(97.68,CNN-RNN GRU)
};
\addlegendentry{Accuracy}
\addlegendimage{/pgfplots/refstyle=EERplotRB,color=black,fill=red}\addlegendentry{EER}
\end{axis}
\end{tikzpicture}
&
\begin{tikzpicture}[scale=0.55, every node/.style={scale=1.0}]
\begin{axis}[
xbar,
axis x line*=bottom,
width  = 0.65*\textwidth,
height = 6.5cm,
xmin=0,xmax=0.32,
xtick={0.00,0.05,0.10,0.15,0.20,0.25,0.30},
xlabel = {EER},
major y tick style = transparent,
xbar=5*\pgflinewidth,
bar width=16.0pt,
bar shift=-9.0pt,
x tick label style={
    	/pgf/number format/.cd,
   	fixed,
   	fixed zerofill,
    	precision=2},
y tick label style={
		font=\footnotesize,
%		anchor=north east,
%		inner sep=0mm
		},
symbolic y coords={%
	CNN-RNN RNN,
	CNN-RNN LSTM,
	CNN-RNN GRU
	},
yticklabels={%
	CNN-RNN,
	CNN-LSTM,
	CNN-GRU
	},
ytick=data,
enlarge y limits=0.275,
%legend cell align=left,
%legend style={
%%	at={(1,1.05)},
%%	anchor=south east,
%%	nodes={rotate=90},%%%%% rotate text in legend
%%	at={(0.125,0)},
%%	at={(0.125,0)},
%%	at={(0.8775,0)},
%	at={(0.225,0.0175)},
%	anchor=south,
%	column sep=1ex
%},
nodes near coords, 
%nodes near coords align={horizontal},
every node near coord/.append style={
	anchor=west,
	font=\footnotesize,
	/pgf/number format/.cd,
	fixed,
	fixed zerofill,
	precision=4},
ytick=data,
]
\addplot [fill=red,opacity=1.00]
coordinates {
(0.1790,CNN-RNN RNN)
(0.1431,CNN-RNN LSTM)
(0.1061,CNN-RNN GRU)
};
%\addlegendentry{EER}
\label{EERplotRC}
\end{axis}
%
%
%%%%%%%%%%%%%%%%%%%%%%%%%%%%%%%%%%%%%%%%%%
%
%
\begin{axis}[ 
xbar,
axis x line*=top,
width  = 0.65*\textwidth,
height = 6.5cm,
xmin=0,xmax=112.5,
xtick={0,20,40,60,80,100},
xlabel = {Accuracy (percentage)},
major y tick style = transparent,
xbar=5*\pgflinewidth,
bar width=16.0pt,
bar shift=9.0pt,
x tick label style={
    	/pgf/number format/.cd,
   	fixed,
   	fixed zerofill,
    	precision=0},
y tick label style={
		font=\footnotesize,
%		anchor=north east,
%		inner sep=0mm
		},
symbolic y coords={%
	CNN-RNN RNN,
	CNN-RNN LSTM,
	CNN-RNN GRU
	},
yticklabels={%
	CNN-RNN,
	CNN-LSTM,
	CNN-GRU
	},
ytick=data,
enlarge y limits=0.275,
legend cell align=left,
legend style={
%	at={(1,1.05)},
%	anchor=south east,
%	nodes={rotate=90},%%%%% rotate text in legend
%	at={(0.125,0)},
%	at={(0.125,0)},
%	at={(0.8775,0)},
%	at={(0.825,0.0175)},
%	at={(0.155,0.825)},
%	at={(0.5,0.825)},
%	at={(0.5,0.675)},
%	at={(0.165,0.05)},
	at={(0.1575,0.03)},
	anchor=south,
	column sep=1ex
},
nodes near coords, 
%nodes near coords align={horizontal},
every node near coord/.append style={
	anchor=west,
	font=\footnotesize,
	/pgf/number format/.cd,
	fixed,
	fixed zerofill,
	precision=2},
ytick=data,
]
%\addplot [fill=blue,opacity=0.10]
%coordinates {
%(99.06,CNN-GRU-without-sampler-fine-tune)
%(98.24,CNN-GRU-with-sampler)
%(98.20,CNN-GRU-without-sampler-not-best-val)
%(97.84,CNN-GRU-without-sampler-best-EER)
%(98.25,CNN-GRU-without-sampler-best-val)
%(98.35,CNN-GRU-cross-entropy-loss)
%(89.89,CNN-GRU-transformer-encoder)
%(94.39,CNN-GRU-pre-trained-word-embedding)
%};
%\end{axis}
%\addlegendentry{EER}
\addplot [fill=blue,opacity=1.00]
coordinates {
(81.65,CNN-RNN RNN)
(85.18,CNN-RNN LSTM)
(88.62,CNN-RNN GRU)
};
\addlegendentry{Accuracy}
\addlegendimage{/pgfplots/refstyle=EERplotRC,color=black,fill=red}\addlegendentry{EER}
\end{axis}
\end{tikzpicture}
\\
(a) Buffalo dataset
&
(b) Clarkson~II dataset
\end{tabular}
\caption{CNN-RNN}\label{fig:5.8-9}
\end{figure}

For the Buffalo keystroke dataset, the performances of our three different models
are virtually identical, which indicates that the most valuable information 
is contained in adjacent keystroke pairs. However, for the Clarkson~II keystroke dataset, 
we find that the GRU is more effective than the other two architectures. 
A plausible explanation is that LSTM
is more prone to overfitting, while RNN is simply less powerful.
And it appears that the GRU is slightly better 
at dealing with noisy data. 

\subsubsection{Cutout Experiments}

\begin{figure}[!htb]
\centering
\begin{tabular}{cc}
\begin{tikzpicture}[scale=0.55, every node/.style={scale=1.0}]
\begin{axis}[
xbar,
axis x line*=bottom,
width  = 0.65*\textwidth,
height = 7.5cm,
xmin=0,xmax=0.32,
xtick={0.00,0.05,0.10,0.15,0.20,0.25,0.30},
xlabel = {EER},
major y tick style = transparent,
xbar=5*\pgflinewidth,
bar width=14.0pt,
bar shift=-8.0pt,
x tick label style={
    	/pgf/number format/.cd,
   	fixed,
   	fixed zerofill,
    	precision=2},
y tick label style={
		font=\footnotesize,
%		anchor=north east,
%		inner sep=0mm
		},
symbolic y coords={%
	CNN-RNN,
	CNN-RNN cutout,
	CNN,
	CNN cutout
	},
ytick=data,
enlarge y limits=0.17,
%legend cell align=left,
%legend style={
%%	at={(1,1.05)},
%%	anchor=south east,
%%	nodes={rotate=90},%%%%% rotate text in legend
%%	at={(0.125,0)},
%%	at={(0.125,0)},
%%	at={(0.8775,0)},
%	at={(0.225,0.0175)},
%	anchor=south,
%	column sep=1ex
%},
nodes near coords, 
%nodes near coords align={horizontal},
every node near coord/.append style={
	anchor=west,
	font=\footnotesize,
	/pgf/number format/.cd,
	fixed,
	fixed zerofill,
	precision=4},
ytick=data,
]
\addplot [fill=red,opacity=1.00]
coordinates {
(0.0189,CNN-RNN)
(0.0142,CNN-RNN cutout)
(0.0133,CNN)
(0.0078,CNN cutout)
};
%\addlegendentry{EER}
\label{EERplotCB}
\end{axis}
%
%
%%%%%%%%%%%%%%%%%%%%%%%%%%%%%%%%%%%%%%%%%%
%
%
\begin{axis}[ 
xbar,
axis x line*=top,
width  = 0.65*\textwidth,
height = 7.5cm,
xmin=0,xmax=112.5,
xtick={0,20,40,60,80,100},
xlabel = {Accuracy (percentage)},
major y tick style = transparent,
xbar=5*\pgflinewidth,
bar width=14.0pt,
bar shift=8.0pt,
x tick label style={
    	/pgf/number format/.cd,
   	fixed,
   	fixed zerofill,
    	precision=0},
y tick label style={
		font=\footnotesize,
%		anchor=north east,
%		inner sep=0mm
		},
symbolic y coords={%
	CNN-RNN,
	CNN-RNN cutout,
	CNN,
	CNN cutout
	},
ytick=data,
enlarge y limits=0.17,
legend cell align=left,
legend style={
%	at={(1,1.05)},
%	anchor=south east,
%	nodes={rotate=90},%%%%% rotate text in legend
%	at={(0.125,0)},
%	at={(0.125,0)},
%	at={(0.8775,0)},
%	at={(0.825,0.0175)},
%	at={(0.155,0.825)},
%	at={(0.5,0.825)},
%	at={(0.5,0.675)},
%	at={(0.165,0.05)},
%	at={(0.1575,0.03)},
	at={(0.5,0.78)},
	anchor=south,
	column sep=1ex
},
nodes near coords, 
%nodes near coords align={horizontal},
every node near coord/.append style={
	anchor=west,
	font=\footnotesize,
	/pgf/number format/.cd,
	fixed,
	fixed zerofill,
	precision=2},
ytick=data,
]
%\addplot [fill=blue,opacity=0.10]
%coordinates {
%(99.06,CNN-GRU-without-sampler-fine-tune)
%(98.24,CNN-GRU-with-sampler)
%(98.20,CNN-GRU-without-sampler-not-best-val)
%(97.84,CNN-GRU-without-sampler-best-EER)
%(98.25,CNN-GRU-without-sampler-best-val)
%(98.35,CNN-GRU-cross-entropy-loss)
%(89.89,CNN-GRU-transformer-encoder)
%(94.39,CNN-GRU-pre-trained-word-embedding)
%};
%\end{axis}
%\addlegendentry{EER}
\addplot [fill=blue,opacity=1.00]
coordinates {
(96.56,CNN-RNN)
(97.68,CNN-RNN cutout)
(97.75,CNN)
(98.53,CNN cutout)
};
\addlegendentry{Accuracy}
\addlegendimage{/pgfplots/refstyle=EERplotCB,color=black,fill=red}\addlegendentry{EER}
\end{axis}
\end{tikzpicture}
&
\begin{tikzpicture}[scale=0.55, every node/.style={scale=1.0}]
\begin{axis}[
xbar,
axis x line*=bottom,
width  = 0.65*\textwidth,
height = 7.5cm,
xmin=0,xmax=0.32,
xtick={0.00,0.05,0.10,0.15,0.20,0.25,0.30},
xlabel = {EER},
major y tick style = transparent,
xbar=5*\pgflinewidth,
bar width=14.0pt,
bar shift=-8.0pt,
x tick label style={
    	/pgf/number format/.cd,
   	fixed,
   	fixed zerofill,
    	precision=2},
y tick label style={
		font=\footnotesize,
%		anchor=north east,
%		inner sep=0mm
		},
symbolic y coords={%
	CNN-RNN,
	CNN-RNN cutout,
	CNN,
	CNN cutout
	},
ytick=data,
enlarge y limits=0.17,
%legend cell align=left,
%legend style={
%%	at={(1,1.05)},
%%	anchor=south east,
%%	nodes={rotate=90},%%%%% rotate text in legend
%%	at={(0.125,0)},
%%	at={(0.125,0)},
%%	at={(0.8775,0)},
%	at={(0.225,0.0175)},
%	anchor=south,
%	column sep=1ex
%},
nodes near coords, 
%nodes near coords align={horizontal},
every node near coord/.append style={
	anchor=west,
	font=\footnotesize,
	/pgf/number format/.cd,
	fixed,
	fixed zerofill,
	precision=4},
ytick=data,
]
\addplot [fill=red,opacity=1.00]
coordinates {
(0.1097,CNN-RNN)
(0.1061,CNN-RNN cutout)
(0.0780,CNN)
(0.0716,CNN cutout)
};
%\addlegendentry{EER}
\label{EERplotCC}
\end{axis}
%
%
%%%%%%%%%%%%%%%%%%%%%%%%%%%%%%%%%%%%%%%%%%
%
%
\begin{axis}[ 
xbar,
axis x line*=top,
width  = 0.65*\textwidth,
height = 7.5cm,
xmin=0,xmax=112.5,
xtick={0,20,40,60,80,100},
xlabel = {Accuracy (percentage)},
major y tick style = transparent,
xbar=5*\pgflinewidth,
bar width=14.0pt,
bar shift=8.0pt,
x tick label style={
    	/pgf/number format/.cd,
   	fixed,
   	fixed zerofill,
    	precision=0},
y tick label style={
		font=\footnotesize,
%		anchor=north east,
%		inner sep=0mm
		},
symbolic y coords={%
	CNN-RNN,
	CNN-RNN cutout,
	CNN,
	CNN cutout
	},
ytick=data,
enlarge y limits=0.17,
legend cell align=left,
legend style={
%	at={(1,1.05)},
%	anchor=south east,
%	nodes={rotate=90},%%%%% rotate text in legend
%	at={(0.125,0)},
%	at={(0.125,0)},
%	at={(0.8775,0)},
%	at={(0.825,0.0175)},
%	at={(0.155,0.825)},
%	at={(0.5,0.825)},
%	at={(0.5,0.675)},
%	at={(0.165,0.05)},
	at={(0.1575,0.03)},
	anchor=south,
	column sep=1ex
},
nodes near coords, 
%nodes near coords align={horizontal},
every node near coord/.append style={
	anchor=west,
	font=\footnotesize,
	/pgf/number format/.cd,
	fixed,
	fixed zerofill,
	precision=2},
ytick=data,
]
%\addplot [fill=blue,opacity=0.10]
%coordinates {
%(99.06,CNN-GRU-without-sampler-fine-tune)
%(98.24,CNN-GRU-with-sampler)
%(98.20,CNN-GRU-without-sampler-not-best-val)
%(97.84,CNN-GRU-without-sampler-best-EER)
%(98.25,CNN-GRU-without-sampler-best-val)
%(98.35,CNN-GRU-cross-entropy-loss)
%(89.89,CNN-GRU-transformer-encoder)
%(94.39,CNN-GRU-pre-trained-word-embedding)
%};
%\end{axis}
%\addlegendentry{EER}
\addplot [fill=blue,opacity=1.00]
coordinates {
(88.33,CNN-RNN)
(88.62,CNN-RNN cutout)
(91.40,CNN)
(92.09,CNN cutout)
};
\addlegendentry{Accuracy}
\addlegendimage{/pgfplots/refstyle=EERplotCC,color=black,fill=red}\addlegendentry{EER}
\end{axis}
\end{tikzpicture}
\\
(a) Buffalo dataset
&
(b) Clarkson~II dataset
\end{tabular}
\caption{Cutout regularization}\label{fig:5.10-11}
\end{figure}
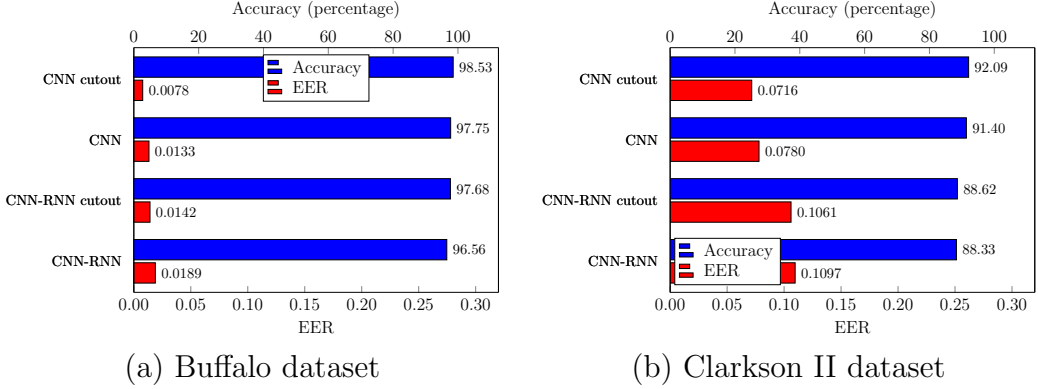

It is likely that the data extracted from keystroke dynamics sequences is 
noisy because of the various extraneous factors that can influence typing behavior. 
We use cutout regularization, since it is useful at preventing overfitting, 
and since it is believed to reduce the effect of noisy information. 
The results of our cutout experiments are given in Figure~\ref{fig:5.10-11}.
We observe that cutout regularization has a significant positive effect on the
performance of our models, which is most obvious in the CNN-based model. 
This is reasonable, since the cutout concept derives from the field of computer vision 
and our input data (i.e., KDI) is an image-like data structure.

\subsection{Discussion}

\begin{figure}[!htb]
\centering
\begin{tabular}{cc}
\begin{tikzpicture}[scale=0.55, every node/.style={scale=1.0}]
\begin{axis}[
xbar,
axis x line*=bottom,
width  = 0.65*\textwidth,
height = 6.5cm,
xmin=0,xmax=0.32,
xtick={0.00,0.05,0.10,0.15,0.20,0.25,0.30},
xlabel = {EER},
major y tick style = transparent,
xbar=5*\pgflinewidth,
bar width=16.0pt,
bar shift=-9.0pt,
x tick label style={
    	/pgf/number format/.cd,
   	fixed,
   	fixed zerofill,
    	precision=2},
y tick label style={
		font=\footnotesize,
%		anchor=north east,
%		inner sep=0mm
		},
symbolic y coords={%
	CNN-RNN one-hot cutout GRU,
	CNN cutout,
	Xiaofeng
	},
yticklabels={%
	CNN-GRU one-hot cutout,
	CNN cutout,
	{Xiaofeng, et al.~\cite{a}}
	},
ytick=data,
enlarge y limits=0.275,
%legend cell align=left,
%legend style={
%%	at={(1,1.05)},
%%	anchor=south east,
%%	nodes={rotate=90},%%%%% rotate text in legend
%%	at={(0.125,0)},
%%	at={(0.125,0)},
%%	at={(0.8775,0)},
%	at={(0.225,0.0175)},
%	anchor=south,
%	column sep=1ex
%},
nodes near coords, 
%nodes near coords align={horizontal},
every node near coord/.append style={
	anchor=west,
	font=\footnotesize,
	/pgf/number format/.cd,
	fixed,
	fixed zerofill,
	precision=4},
]
\addplot [fill=red,opacity=1.00]
coordinates {
(0.0085,CNN-RNN one-hot cutout GRU)
(0.0088,CNN cutout)
(0.0267,Xiaofeng)
};
%\addlegendentry{EER}
\label{EERplotBB}
\end{axis}
%
%
%%%%%%%%%%%%%%%%%%%%%%%%%%%%%%%%%%%%%%%%%%
%
%
\begin{axis}[ 
xbar,
axis x line*=top,
width  = 0.65*\textwidth,
height = 6.5cm,
xmin=0,xmax=112.5,
xtick={0,20,40,60,80,100},
xlabel = {Accuracy (percentage)},
major y tick style = transparent,
xbar=5*\pgflinewidth,
bar width=16.0pt,
bar shift=9.0pt,
x tick label style={
    	/pgf/number format/.cd,
   	fixed,
   	fixed zerofill,
    	precision=0},
y tick label style={
		font=\footnotesize,
%		anchor=north east,
%		inner sep=0mm
		},
symbolic y coords={%
	CNN-RNN one-hot cutout GRU,
	CNN cutout,
	Xiaofeng
	},
yticklabels={%
	CNN-GRU one-hot cutout,
	CNN cutout,
	{Xiaofeng, et al.~\cite{a}}
	},
ytick=data,
enlarge y limits=0.275,
legend cell align=left,
legend style={
%	at={(1,1.05)},
%	anchor=south east,
%	nodes={rotate=90},%%%%% rotate text in legend
%	at={(0.125,0)},
%	at={(0.125,0)},
%	at={(0.8775,0)},
%	at={(0.825,0.0175)},
%	at={(0.155,0.825)},
%	at={(0.5,0.825)},
%	at={(0.5,0.675)},
%	at={(0.165,0.05)},
%	at={(0.1575,0.03)},
	at={(0.5,0.74)},
	anchor=south,
	column sep=1ex
},
nodes near coords, 
%nodes near coords align={horizontal},
every node near coord/.append style={
	anchor=west,
	font=\footnotesize,
	/pgf/number format/.cd,
	fixed,
	fixed zerofill,
	precision=2},
]
%\addplot [fill=blue,opacity=0.10]
%coordinates {
%(99.06,CNN-GRU-without-sampler-fine-tune)
%(98.24,CNN-GRU-with-sampler)
%(98.20,CNN-GRU-without-sampler-not-best-val)
%(97.84,CNN-GRU-without-sampler-best-EER)
%(98.25,CNN-GRU-without-sampler-best-val)
%(98.35,CNN-GRU-cross-entropy-loss)
%(89.89,CNN-GRU-transformer-encoder)
%(94.39,CNN-GRU-pre-trained-word-embedding)
%};
%\end{axis}
%\addlegendentry{EER}
\addplot [fill=blue,opacity=1.00]
coordinates {
(98.50,CNN-RNN one-hot cutout GRU)
(98.56,CNN cutout)
};
\addlegendentry{Accuracy}
\addlegendimage{/pgfplots/refstyle=EERplotBB,color=black,fill=red}\addlegendentry{EER}
\addplot [fill=blue,opacity=1.00,nodes near coords = {N\kern-1pt \texttt{/}\kern-2pt A}]
coordinates {
(0.0,Xiaofeng)
};
%\addlegendentry{Accuracy}
%\addlegendimage{/pgfplots/refstyle=EERplot,color=black,fill=red}\addlegendentry{EER}
\end{axis}
\end{tikzpicture}
&
\begin{tikzpicture}[scale=0.55, every node/.style={scale=1.0}]
\begin{axis}[
xbar,
axis x line*=bottom,
width  = 0.65*\textwidth,
height = 6.5cm,
xmin=0,xmax=0.32,
xtick={0.00,0.05,0.10,0.15,0.20,0.25,0.30},
xlabel = {EER},
major y tick style = transparent,
xbar=5*\pgflinewidth,
bar width=16.0pt,
bar shift=-9.0pt,
x tick label style={
    	/pgf/number format/.cd,
   	fixed,
   	fixed zerofill,
    	precision=2},
y tick label style={
		font=\footnotesize,
%		anchor=north east,
%		inner sep=0mm
		},
symbolic y coords={%
	CNN-RNN one-hot cutout GRU,
	CNN cutout,
	Xiaofeng
	},
yticklabels={%
	CNN-GRU one-hot cutout,
	CNN cutout,
	{Xiaofeng, et al.~\cite{a}}
	},
ytick=data,
enlarge y limits=0.275,
%legend cell align=left,
%legend style={
%%	at={(1,1.05)},
%%	anchor=south east,
%%	nodes={rotate=90},%%%%% rotate text in legend
%%	at={(0.125,0)},
%%	at={(0.125,0)},
%%	at={(0.8775,0)},
%	at={(0.225,0.0175)},
%	anchor=south,
%	column sep=1ex
%},
nodes near coords, 
%nodes near coords align={horizontal},
every node near coord/.append style={
	anchor=west,
	font=\footnotesize,
	/pgf/number format/.cd,
	fixed,
	fixed zerofill,
	precision=4},
]
\addplot [fill=red,opacity=1.00]
coordinates {
(0.0774,CNN-RNN one-hot cutout GRU)
(0.0755,CNN cutout)
(0.0597,Xiaofeng)
};
%\addlegendentry{EER}
\label{EERplotBC}
\end{axis}
%
%
%%%%%%%%%%%%%%%%%%%%%%%%%%%%%%%%%%%%%%%%%%
%
%
\begin{axis}[ 
xbar,
axis x line*=top,
width  = 0.65*\textwidth,
height = 6.5cm,
xmin=0,xmax=112.5,
xtick={0,20,40,60,80,100},
xlabel = {Accuracy (percentage)},
major y tick style = transparent,
xbar=5*\pgflinewidth,
bar width=16.0pt,
bar shift=9.0pt,
x tick label style={
    	/pgf/number format/.cd,
   	fixed,
   	fixed zerofill,
    	precision=0},
y tick label style={
		font=\footnotesize,
%		anchor=north east,
%		inner sep=0mm
		},
symbolic y coords={%
	CNN-RNN one-hot cutout GRU,
	CNN cutout,
	Xiaofeng
	},
yticklabels={%
	CNN-GRU one-hot cutout,
	CNN cutout,
	{Xiaofeng, et al.~\cite{a}}
	},
ytick=data,
enlarge y limits=0.275,
legend cell align=left,
legend style={
%	at={(1,1.05)},
%	anchor=south east,
%	nodes={rotate=90},%%%%% rotate text in legend
%	at={(0.125,0)},
%	at={(0.125,0)},
%	at={(0.8775,0)},
%	at={(0.825,0.0175)},
%	at={(0.155,0.825)},
%	at={(0.5,0.825)},
%	at={(0.5,0.675)},
%	at={(0.165,0.05)},
%	at={(0.1575,0.03)},
	at={(0.5,0.74)},
	anchor=south,
	column sep=1ex
},
nodes near coords, 
%nodes near coords align={horizontal},
every node near coord/.append style={
	anchor=west,
	font=\footnotesize,
	/pgf/number format/.cd,
	fixed,
	fixed zerofill,
	precision=2},
]
%\addplot [fill=blue,opacity=0.10]
%coordinates {
%(99.06,CNN-GRU-without-sampler-fine-tune)
%(98.24,CNN-GRU-with-sampler)
%(98.20,CNN-GRU-without-sampler-not-best-val)
%(97.84,CNN-GRU-without-sampler-best-EER)
%(98.25,CNN-GRU-without-sampler-best-val)
%(98.35,CNN-GRU-cross-entropy-loss)
%(89.89,CNN-GRU-transformer-encoder)
%(94.39,CNN-GRU-pre-trained-word-embedding)
%};
%\end{axis}
%\addlegendentry{EER}
\addplot [fill=blue,opacity=1.00]
coordinates {
(91.91,CNN-RNN one-hot cutout GRU)
(91.74,CNN cutout)
};
\addlegendentry{Accuracy}
\addlegendimage{/pgfplots/refstyle=EERplotBC,color=black,fill=red}\addlegendentry{EER}
\addplot [fill=blue,opacity=1.00,nodes near coords = {N\kern-1pt \texttt{/}\kern-2pt A}]
coordinates {
(0.0,Xiaofeng)
};
%\addlegendentry{Accuracy}
%\addlegendimage{/pgfplots/refstyle=EERplot,color=black,fill=red}\addlegendentry{EER}
\end{axis}
\end{tikzpicture}
\\
(a) Buffalo dataset
&
(b) Clarkson~II dataset
\end{tabular}
\caption{Best results and comparison to previous work}\label{fig:5.12}
\end{figure}
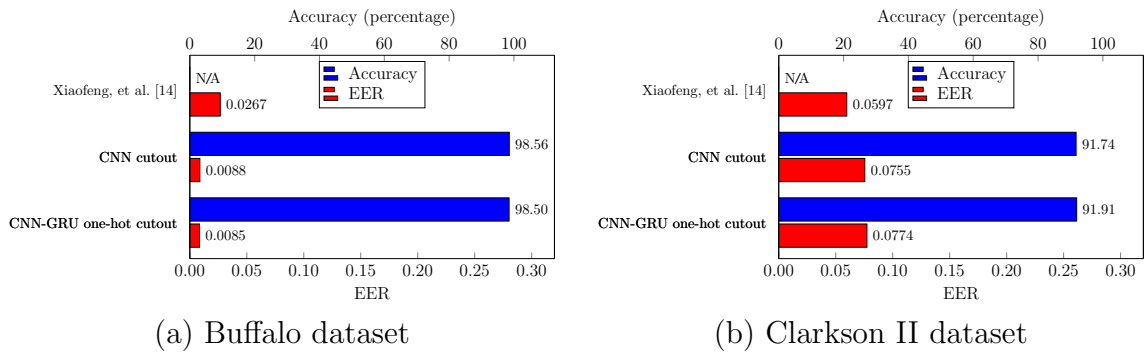

In our experiments, the performance on the Buffalo dataset is 
consistently higher than that of the Clarkson~II dataset. It is likely the case that the 
latter dataset contains noisier data, as it was collected 
over a period of~2.5 years and under far less controlled conditions. 
We also find that our CNN-based model ($\mbox{KDI}+\mbox{CNN}$) consistently generates
better results than our RNN-CNN based model ($\mbox{Free-KDS}+\mbox{CNN-RNN}$).
Comparing our results with the previous work in~\cite{a}, we
observe that in terms of EER, our two models both perform better on the Buffalo dataset, 
but slightly worse on the Clarkson~II dataset. These results are summarized
in Figure~\ref{fig:5.12}.

\section{Conclusion}\label{chap:conclusion}

This research focused on authentication based on keystroke dynamics 
derived features in the free-text case. We found that 
dividing the sequence into a number of fixed-length subsequences was an effective
feature engineering strategy.
In addition, we developed and analyzed an image-like engineered
feature structure that we refer to as~KDI, 
and we compared this to another structure that we refer to as~KDS.
The KDI was used as the input for our CNN experiments, 
while the KDS served as the input data for our CNN-RNN experiments. 
In both cases, we applied cutout regularization. 

The experimental results reported here show that our pure CNN architecture outperforms
our combination of CNN and RNN, and cutout significantly improves the performances 
of both models. Moreover, our two modeling approaches both outperform previous work on
the Buffalo keystroke dataset and yield competitive results for the Clarkson~II dataset.

In the realm of future work, we conjecture that generative adversarial
networks (GAN) will prove useful in this problem domain.
More fundamentally, we believe that improved (and larger) datasets are 
necessary if we are to make significant further progress on this 
challenging authentication problem.

\bibliographystyle{plain}
\bibliography{referencesJW.bib}

\begin{thebibliography}{10}

\bibitem{9}
Saad Albawi, Tareq~Abed Mohammed, and Saad Al-Zawi.
\newblock Understanding of a convolutional neural network.
\newblock In {\em 2017 International Conference on Engineering and Technology},
  ICET, pages 1--6, 2017.

\bibitem{2}
Faisal Alshanketi, Issa Traore, and Ahmed~Awad Ahmed.
\newblock Improving performance and usability in mobile keystroke dynamic
  biometric authentication.
\newblock In {\em 2016 IEEE Security and Privacy Workshops}, SPW, pages 66--73,
  2016.

\bibitem{d}
Mario~Luca Bernardi, Marta Cimitile, Fabio Martinelli, and Francesco Mercaldo.
\newblock Keystroke analysis for user identification using deep neural
  networks.
\newblock In {\em 2019 International Joint Conference on Neural Networks},
  IJCNN, pages 1--8, 2019.

\bibitem{14}
Hayreddin \c{C}eker and Shambhu Upadhyaya.
\newblock Enhanced recognition of keystroke dynamics using gaussian mixture
  models.
\newblock In {\em 2015 IEEE Military Communications Conference}, MILCOM, pages
  1305--1310, 2015.

\bibitem{15}
Hayreddin \c{C}eker and Shambhu Upadhyaya.
\newblock Sensitivity analysis in keystroke dynamics using convolutional neural
  networks.
\newblock In {\em 2017 IEEE Workshop on Information Forensics and Security},
  WIFS, pages 1--6, 2017.

\bibitem{19}
Junyoung Chung, Caglar Gulcehre, KyungHyun Cho, and Yoshua Bengio.
\newblock Empirical evaluation of gated recurrent neural networks on sequence
  modeling.
\newblock \url{https://arxiv.org/abs/1412.3555}, 2014.

\bibitem{11}
Terrance DeVries and Graham~W Taylor.
\newblock Improved regularization of convolutional neural networks with cutout.
\newblock \url{https://arxiv.org/abs/1708.04552}, 2017.

\bibitem{10}
Sepp Hochreiter and J{\"u}rgen Schmidhuber.
\newblock Long short-term memory.
\newblock {\em Neural Computation}, 9(8):1735--1780, 1997.

\bibitem{12}
Pilsung Kang and Sungzoon Cho.
\newblock Keystroke dynamics-based user authentication using long and free text
  strings from various input devices.
\newblock {\em Information Sciences}, 308:72--93, 2015.

\bibitem{b}
Junhong Kim, Haedong Kim, and Pilsung Kang.
\newblock Keystroke dynamics-based user authentication using freely typed text
  based on user-adaptive feature extraction and novelty detection.
\newblock {\em Applied Soft Computing}, 62:1077--1087, 2018.

\bibitem{u}
Paweł Kobojek and Khalid Saeed.
\newblock Application of recurrent neural networks for user verification based
  on keystroke dynamics.
\newblock {\em Journal of Telecommunications and Information Technology},
  2016:80--90, 2016.

\bibitem{e}
Gutha~Jaya Krishna, Harshal Jaiswal, P.~Sai~Ravi Teja, and Vadlamani Ravi.
\newblock Keystroke based user identification with {XGBoost}.
\newblock In {\em 2019 IEEE Region 10 Conference}, TENCON, pages 1369--1374,
  2019.

\bibitem{z}
Andreas Lanitis.
\newblock A survey of the effects of aging on biometric identity verification.
\newblock {\em International Journal of Biometrics}, 2(1):34–52, 2010.

\bibitem{a}
Xiaofeng Lu, Shengfei Zhang, and Shengwei Yi.
\newblock Free-text keystroke continuous authentication using {CNN} and {RNN}.
\newblock {\em Journal of Tsinghua University (Science and Technology)},
  58(12):1072--1078, 2018.

\bibitem{0}
Andrew Maas, Chris Heather, Chuong~(Tom) Do, Relly Brandman, Daphne Koller, and
  Andrew Ng.
\newblock Offering verified credentials in massive open online courses: {MOOC}s
  and technology to advance learning and learning research.
\newblock {\em Ubiquity}, 2014:1--11, 2014.

\bibitem{16}
Fabian Monrose and Aviel~D. Rubin.
\newblock Keystroke dynamics as a biometric for authentication.
\newblock {\em Future Generation Computer Systems}, 16(4):351--359, 2000.

\bibitem{f}
Eduard~C. Popovici, Ovidiu~G. Guta, Liviu~A. Stancu, Stefan~C. Arseni, and
  Octavian Fratu.
\newblock {MLP} neural network for keystroke-based user identification system.
\newblock In {\em 11th International Conference on Telecommunications in Modern
  Satellite, Cable and Broadcasting Services}, TELSIKS, pages 155--158, 2013.

\bibitem{8}
J{\"u}rgen Schmidhuber.
\newblock Deep learning in neural networks: An overview.
\newblock {\em Neural Networks}, 61:85--117, 2015.

\bibitem{17}
Nitish Srivastava, Geoffrey~E. Hinton, Alex Krizhevsky, Ilya Sutskever, and
  Ruslan Salakhutdinov.
\newblock Dropout: A simple way to prevent neural networks from overfitting.
\newblock {\em Journal of Machine Learning Research}, 15(56):1929--1958, 2014.

\bibitem{i}
Y.~Sun, Hayreddin \c{C}eker, and Shambhu Upadhyaya.
\newblock Shared keystroke dataset for continuous authentication.
\newblock In {\em 2016 IEEE International Workshop on Information Forensics and
  Security}, WIFS, pages 1--6, 2016.

\bibitem{clark}
Esra Vural, Jiaju Huang, Daqing Hou, and Stephanie Schuckers.
\newblock {Clarkson University} keystroke dataset.
\newblock
  \url{https://citer.clarkson.edu/research-resources/biometric-dataset-collections-2/clarkson-university-keystroke-dataset/}.

\bibitem{1}
Jatin Yadav, Kavita Pandey, Shashank Gupta, and Richa Sharma.
\newblock Keystroke dynamics based authentication using fuzzy logic.
\newblock In {\em 2017 Tenth International Conference on Contemporary
  Computing}, IC3, pages 1--6, 2017.

\end{thebibliography}

\end{document}